\documentclass[journal]{IEEEtran}

\usepackage{times}
\usepackage{epsfig}
\usepackage{graphicx}
\usepackage{subcaption}
\usepackage{amsmath}
\usepackage{amssymb}
\usepackage{color}
\usepackage{multirow}
\usepackage{bm}
\usepackage{dsfont}
\usepackage{algorithm}
\usepackage{algorithmic}
\usepackage{array}
\usepackage{textcomp}
\usepackage[normalem]{ulem}
\usepackage{etoolbox}
\usepackage{xcolor}
\usepackage[utf8x]{inputenc}
\graphicspath{{}}

\usepackage[breaklinks=true,bookmarks=false]{hyperref}

\newcommand{\qed}{\nobreak \ifvmode \relax \else
      \ifdim\lastskip<1.5em \hskip-\lastskip
      \hskip1.5em plus0em minus0.5em \fi \nobreak
      \vrule height0.75em width0.5em depth0.25em\fi}

\usepackage{xspace}

\newcommand{\MyMapTemplatePrefixc}[4]{\expandafter#1\csname#3#4\endcsname{#2{#4}}} 
\forcsvlist{\MyMapTemplatePrefixc {\def} {\mathcal}{c}} {A,B,C,D,E,F,G,H,I,J,K,L,M,N,O,P,Q,R,S,T,U,V,W,X,Y,Z}  

\newcommand{\MyMapTemplatePrefixtb}[5]{\expandafter#1\csname#4#5\endcsname{#2{#3{#5}}}} 
\forcsvlist{\MyMapTemplatePrefixtb {\def} {\tilde}{\mathbf}{t}} {a,b,c,d,e,f,g,h,i,j,k,l,m,n,o,p,q,r,s,t,u,v,w,x,y,z}  

\newcommand{\MyMapTemplateNoPrefix}[3]{\expandafter#1\csname#3\endcsname{#2{#3}}}
\forcsvlist{\MyMapTemplateNoPrefix {\def} {\mathbf}} {0, 1, a, b, c, d, e, f, g, h, i, j, k, l, m, n, o, p, q, r, u, v, w, x, y, z} 
\forcsvlist{\MyMapTemplateNoPrefix {\def} {\mathbf}} {A,B,C,D,E,F,G,H,I,J,K,L,M,N,O,P,Q,R,S,T,U,V,W,X,Y,Z}  

\def\ie{\emph{i.e.}}
\def\eg{\emph{e.g.}}

\def\bbR{{\mathbb R}}

\def\etal{\emph{et al.}\@\xspace}

\def\ie{\emph{i.e.}\@\xspace}
\def\eg{\emph{e.g.}\@\xspace}
\def\resp{\emph{resp.}\@\xspace}
\def\wrt{\emph{w.r.t.}\@\xspace}

\begin{document}
\title{Fine-grained Classification using Heterogeneous Web Data and Auxiliary Categories}

\author{Li Niu, Ashok Veeraraghavan, and Ashu Sabharwal\
\thanks{L. Niu is with Electric and Computer Engineering (ECE) department in Rice University, Houston, TX 77005, USA (e-mail:ln7@rice.edu).}
\thanks{A. Veeraraghavan is with Electric and Computer Engineering (ECE) department in Rice University, Houston, TX 77005, USA (e-mail:vashok@rice.edu).}
\thanks{A. Sabharwal is with Electric and Computer Engineering (ECE) department in Rice University, Houston, TX 77005, USA (e-mail:ashu@rice.edu).}
}

\maketitle

\begin{abstract}
Fine-grained classification, which aims to distinguish the subtle difference among various fine-grained categories belonging to one coarse-grained category, remains a very challenging problem, because of the absence of well-labeled training data caused by the high cost of annotating a large number of fine-grained categories. In the extreme case, given a set of test categories without any well-labeled training data, the majority of existing works can be grouped into the following two research directions: 1) crawl noisy labeled web data for the test categories as training data, which is dubbed as webly supervised learning; 2) transfer the knowledge from auxiliary categories with well-labeled training data to the test categories, which corresponds to zero-shot learning setting. Nevertheless, the above two research directions still have critical issues to be addressed. For the first direction, web data have noisy labels and considerably different data distribution from test data. For the second direction, zero-shot learning is struggling to achieve compelling results compared with conventional supervised learning. The issues of the above two directions motivate us to develop a novel approach which can jointly exploit both noisy web training data from test categories and well-labeled training data from auxiliary categories. In particular, on one hand, we crawl web data for test categories as noisy training data. On the other hand, we transfer the knowledge from auxiliary categories with well-labeled training data to test categories by virtue of free semantic information (\eg, word vector) of all categories. Moreover, given the fact that web data are generally associated with additional textual information (\eg, title and tag), we extend our method by using the surrounding textual information of web data as privileged information. Extensive experiments show the effectiveness of our proposed methods. 
\end{abstract}

\section{Introduction} \label{sec:intro}
Recently, the field of image classification is greatly fueled by the rapid development in deep learning techniques and large-scale image datasets such as ImageNet~\cite{deng2009imagenet}. However, fine-grained image classification, which targets at classifying abundant fine-grained categories belonging to one coarse-grained category (\eg, bird species and dog breeds), is still a very tough task. To identify the minor distinction among various fine-grained categories, sufficient well-labeled training images are in high demand. However, accurate human annotation for fine-grained categories is not easy to acquire due to the following reasons: 1) fine-grained annotation generally requires expertise, which raises the bar for human annotators; 2) there are usually myriads of fine-grained categories belonging to one coarse-grained category (\eg, in total $14,000$ known bird species~\cite{krause2016unreasonable}) and hence it is infeasible to collect well-labeled training images for all fine-grained categories exhaustively. Therefore, the absence of well-labeled training images is a vital issue for fine-grained classification. In this work, we take an extreme case into consideration, that is, there are not any well-labeled training images for a given set of test categories. In this circumstance, the existing research works mainly fall into two research realms, \ie, Webly Supervised Learning (WSL) and Zero-Shot Learning (ZSL)~\cite{lampert2014attribute}, which will be elaborated separately in the following.

For Webly Supervised Learning (WSL), freely available web images are crawled from public websites (\eg, Flickr and Google) using category names as queries. Nevertheless, web images are loosely labeled, which means that their labels are very noisy and often inaccurate~\cite{chen2015webly,zhuang2016attend}. When the classifier is learnt based on noisy training images, its performance on the test set will be significantly degraded. Moreover, when images are uploaded to public websites, they are often edited or compressed, leading to the dramatic data distribution mismatch between web images and test images, which is also referred to as domain shift~\cite{torralba2011unbiased}. Up to date, some works~\cite{krause2016unreasonable,xu2015augmenting} have been proposed for fine-grained classification based on web data, which tend to address the above two issues: label noise and domain shift. However, they rely on strong supervision (\eg, bounding box and part location) on web images or human intervention when collecting web images, which is often inaccessible in the real-world applications. 

For Zero-Shot Learning (ZSL)~\cite{lampert2014attribute}, training categories (\ie, seen categories) and test categories (\ie, unseen categories) have no overlap. In other words, with training instances from seen categories, 
we need to recognize the test instances from unseen categories. To achieve this goal, intermediate category-level semantic information of all categories is used to bridge the gap between seen categories and unseen categories. There are various forms of category-level semantic information including attributes~\cite{lampert2014attribute} (\eg, shape, color, and material), which are manually designed by human experts, and word vectors~\cite{mikolov2013distributed,pennington2014glove} corresponding to category names, which can be obtained via free online corpus (\eg, Wikipedia)~\cite{akata2015evaluation}. However, the performance of ZSL is still far below that of conventional supervised learning~\cite{akata2015evaluation}, especially when using free semantic information such as word vector.

To this end, we propose a new learning scenario for fine-grained classification, which unifies webly supervised learning and zero-shot learning. In particular, given a set of fine-grained categories without any well-labeled training images, we crawl noisy web images for these categories as training data and also utilize the well-labeled images from auxiliary fine-grained categories. From another point of view, given the entire set of fine-grained categories belonging to one coarse-grained category, we only need to ask human experts to annotate partial fine-grained categories, and then can predict the rest of fine-grained categories with the aid of web data.
Therefore, our proposed learning scenario could be viewed as webly supervised learning with well-labeled data from auxiliary categories, or zero-shot learning with noisy web training data for unseen categories. 

We develop our method in this learning scenario. Given a set of auxiliary categories and a set of test categories, we first crawl noisy web training images for test categories using the category names as queries. Then, deep visual features are extracted for all images including web training images, well-labeled training images, and unlabeled test images. Furthermore, intermediate category-level semantic representations are extracted for all categories. In practice, we use word vector~\cite{mikolov2013distributed,pennington2014glove} as category-level semantic representation. To be exact, we train a linguistic model based on free online corpus (\eg, Wikipedia) and obtain the word vector corresponding to each category name. Finally, visual features and word vectors are fed into our learning model, yielding the prediction results of test images. 

Another benefit of learning from web data is that web images are generally associated with additional textual information such as titles and tags, while test images do not have such textual information. The information which is only available for training instances but not available for test instances is dubbed as privileged information~\cite{LUPIparadigm}. The privileged information can be utilized in the training stage to help learn a more robust model, so we take advantage of additional textual information as privileged information in our method. Particularly, we extract textual features from the surrounding textual information of web images, which are fed into our learning model together with visual features and word vectors.

\begin{figure}[t]
	\centering
    \includegraphics[width=0.495\textwidth]{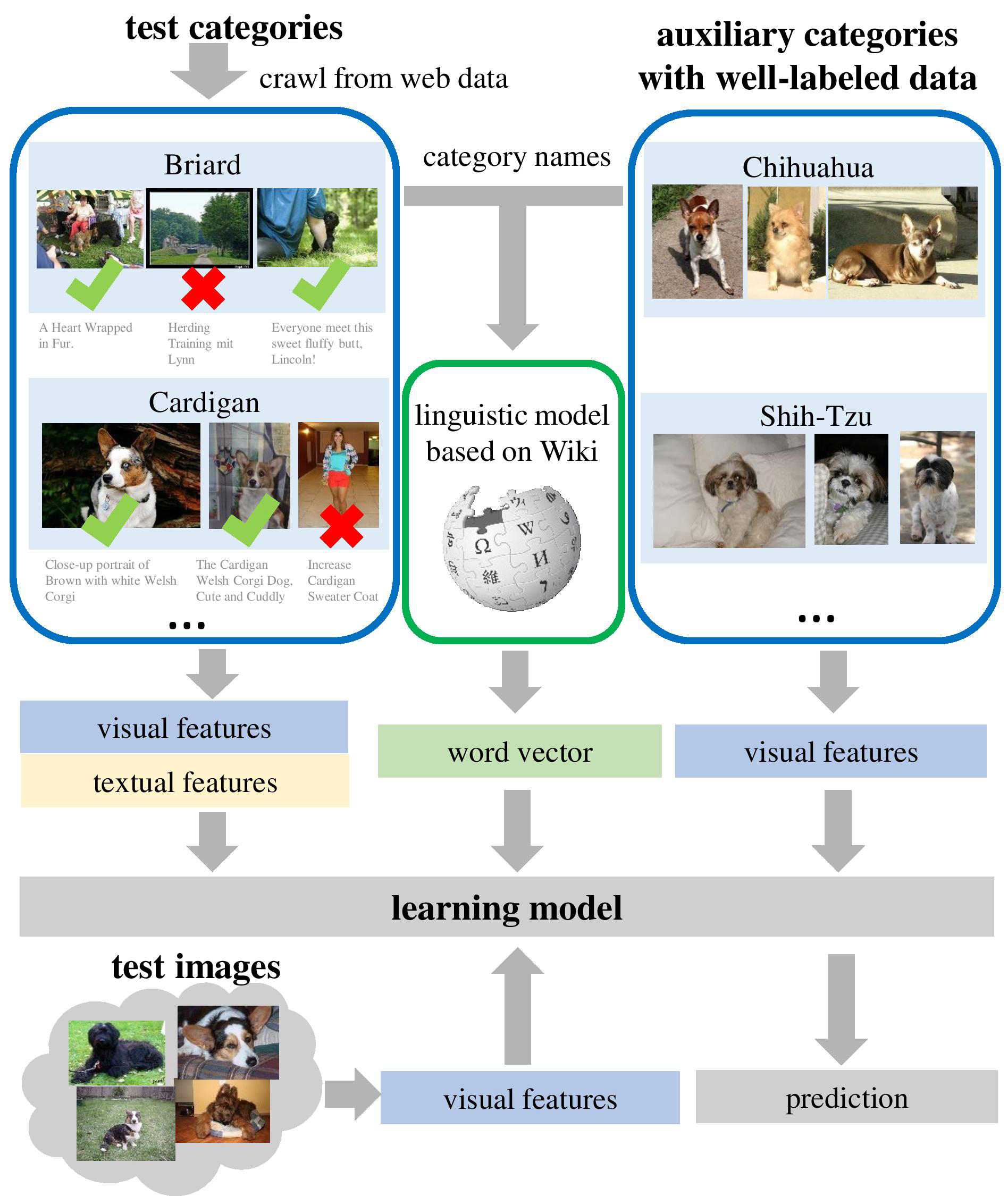}		
    \caption{The flowchart of our method which can utilize both web data (\ie, web images with surrounding textual information and online corpus) and auxiliary categories for fine-grained image classification. We use dog breeds (\ie, Briard, Cardigan, Chihuahua, and Shih-Tzu) as the examples of fine-grained categories.}
    \label{fig:flowchart}
\end{figure}

The flowchart of our method is illustrated in Fig.~\ref{fig:flowchart}. With visual features, textual features, and word vectors as input, our learning model can cope with the label noise and domain shift of web images, transfer the knowledge from auxiliary categories to test categories, and simultaneously take advantage of the privileged information.  It is obvious that our method involves different types of web data with different functions: 1) web images are crawled from public websites (\eg, Flickr and Google) for test categories as noisy training data; 2) the surrounding textual information of web images are used as privileged information; 3) free online corpus (\eg, Wikipedia) is used to obtain semantic representations of all categories for the sake of filling in the gap between test categories and auxiliary categories. The details of our learning model will be fully introduced in Section~\ref{sec:our_formulation}.

Our major contributions are fourfold: 1) as far as we are concerned, we propose the first learning scenario for fine-grained classification using both web data and auxiliary categories; 2) in this learning scenario, we propose a novel method  unifying zero-shot learning and webly supervised learning, which can transfer the knowledge from auxiliary categories to test categories and simultaneously handle the label noise and domain shift of web data; 3) we further extend our method by using the surrounding textual information of web images as privileged information; 4) the effectiveness of our methods is verified by comprehensive experiments on three benchmark datasets.

This paper extends the preliminary conference version~\cite{niu2018webly} in the following ways. From the theoretical aspect, we extend our model by using additional textual information as privileged information in Section~\ref{sec:method_PI}, followed by an effective solution in Appendix~\ref{sec:appendix_WSZSL_PI}. From the experimental aspect, we evaluate the methods using privileged information in Section~\ref{sec:exp_PI}, and also provide more quantitative and qualitative analyses in Section~\ref{sec:exp_wo_PI}.

\setlength{\textfloatsep}{2pt}

\section{Related Work}
In this section, we will discuss some recent works on webly supervised learning and zero-shot learning. Moreover, as  domain shift needs to be addressed when learning from web data, domain adaptation will also be briefly introduced. Finally, we will describe previous works using privileged information, since additional textual information is used as privileged information in our method.

\noindent\textbf{Webly Supervised Learning}: Learning from web data~\cite{chen2013neil,li2014exploiting,niu2016exploiting,xiao2015learning,chen2015webly,niu2017action,niu2018wsci}, which is also dubbed as webly supervised learning, has attracted more and more research interest from the computer vision community. Lots of works have been done to cope with label noise and domain shift when learning from web data. As a pioneering work, NEIL in~\cite{chen2013neil} uses Multi-Instance Learning (MIL) to suppress the label noise of web images. Besides, a couple of domain adaptation methods were studied in~\cite{bergamo2010exploiting} while a weakly supervised domain generalization method was developed in~\cite{niu2015visual}. With the advance of deep learning techniques, several CNN approaches also emerged for learning from web images~\cite{xiao2015learning,sukhbaatar2014training,chen2015webly,divvala2014learning,
xu2016few,zhuang2016attend}. Nevertheless, fine-grained classification is out of focus of the above works.

With regards to exploiting web data for fine-grained classification, label noise is mitigated in~\cite{krause2016unreasonable} via active learning, which in fact involves human intervention. In~\cite{xu2015augmenting}, bounding box annotations are needed for web data. The work in~\cite{van2015building} leverages Flickr images to learn bird classifiers, whereas human annotators take part in the dataset collection.
In a more recent work~\cite{xu2016webly}, label noise and domain shift are handled when learning from web images, but bounding boxes and part landmarks are required, which are unavailable in our application.
Distinguished from all the above works, we focus on webly supervised fine-grained classification without human intervention when collecting web data or strong supervision (\eg, bounding box and part location) on web data.

\noindent\textbf{Zero-Shot Learning}:
Recently, many Zero-Shot Learning (ZSL) methods have been proposed~\cite{romera2015embarrassingly,akata2015evaluation,lampert2014attribute,NiuCVZ19}. Moreover, 
several works~\cite{xu2017transductive,kodirov2015unsupervised,li2015semi} found that it is helpful to use unlabeled test instances from unseen categories in the training process, leading to semi-supervised/transductive ZSL. Inspired by semi-supervised/transductive ZSL, we also utilize unlabeled test instances in the training stage. For more details of existing ZSL approaches, please refer to two recent surveys~\cite{xian2017zero,FuXJXSG18}.
However, all the above ZSL methods ignore the large amount of freely available web images, which could be used for fine-grained classification. In contrast, we propose a new learning scenario, which treats seen categories as auxiliary categories and additionally exploit web images for test categories. 

\noindent\textbf{Domain Adaptation}: Domain adaptation targets at addressing the domain shift between training set (\ie, source domain) and test set (\ie, target domain) by reducing the domain distribution mismatch between two domains. Analogous to semi-supervised/transductive ZSL, for domain adaptation, unlabeled test instances are also used in the training stage.
Existing domain adaptation approaches can be roughly grouped into reweight based method~\cite{Huang2007KMM}, subspace based method~\cite{Gong2012gfk,SA,Pan2011TCA,sun2016return,niu2016domain}, and generative model based method~\cite{BousmalisSDEK17,ZhuPIE17}. Recently, many domain generalization methods~\cite{xu2014exploiting,niu2015multi,niu2017visual,niu2016exemplar} have been proposed. Different from domain adaptation, domain generalization focuses on the setting in which the unlabeled test instances are unavailable in the training stage.
The closest related work is the reweight based method~\cite{Huang2007KMM}, which assigns different weights on different training instances on the premise of Maximum Mean Discrepancy(MMD). However, the work in~\cite{Huang2007KMM} only focuses on domain shift and does not favor our application.

\noindent\textbf{Learning using Privileged Information}: Moreover, our work is relevant to learning using privileged information (LUPI) \cite{LUPI}, in which training instances are associated with additional information (\ie, privileged information) that are not available for test instances. The key idea proposed in~\cite{LUPI} is using privileged information to model the classification difficulty of training instances, which has been generalized to a wide range of applications such as ranking~\cite{RankTransfer}, clustering~\cite{FeyereislA12}, metric learning~\cite{xu2015distance}, Gaussian process~\cite{Hernandez-LobatoSKLQ14}, hashing~\cite{ZhouXPTQG16}, and multi-instance learning~\cite{YangZCO17}. However, all these works do not use privileged information in webly supervised learning. The closest related works are~\cite{li2014exploiting,niu2015visual}, which take advantage of textual information as privileged information when learning from web data. However, they do not borrow the idea of auxiliary categories from zero-shot learning. In contrast, we integrate privileged information into our method, which unifies webly supervised learning and zero-shot learning.

\section{Our Learning Model} \label{sec:our_formulation}
Recall that the flowchart of our method has been illustrated in Fig.~\ref{fig:flowchart} and described in Section~\ref{sec:intro}, which will be recapped in the following. Given a set of auxiliary categories with well-labeled training images and a set of test categories, we crawl web training images with their surrounding textual information for test categories and obtain semantic representations (\ie, word vectors~\cite{mikolov2013distributed,pennington2014glove}) for all categories. Then, we feed the visual features of all training images (\ie, well-labeled training images and web training images), the textual features of additional textual information, and word vectors of all categories into our learning model. Besides, we also feed unlabeled test images into our learning model in the training phase, which can help address domain shift and facilitate zero-shot learning~\cite{xu2017transductive,kodirov2015unsupervised,li2015semi}, resulting in a semi-supervised learning model. The output of our learning model is the predicted semantic representations of test images, which are used for final categorization. Next, we will introduce how to transfer the knowledge from auxiliary categories to test categories, how to exploit web images from test categories, and how to take advantage of the additional textual information as privileged information.

In the remainder of this paper, for ease of representation, a matrix/vector is denoted by an uppercase/lowercase letter in boldface. $\A^{-1}$ is used to denote the inverse matrix of $\A$. Moreover, we adopt superscript~$'$ to represent the transpose of a vector/matrix.  We use $\O$ (\resp, $\I$) to denote the zero matrix (\resp, identify matrix). Similarly, we use $\1$ to denote the vector with all ones. $\A\circ\B$ (\resp, $\langle\A,\B\rangle$) is used to denote the element-wise product (\resp, inner product) of two matrices.

\subsection{Knowledge Transfer from Auxiliary Categories to Test Categories} \label{sec:method_knowledge_transfer}

We use $\X^a\in\mathcal{R}^{d\times n^a}$ to denote the visual features of well-labeled training images from $C^a$ auxiliary categories, in which $n^a$ is the number of training images and $d$ is the dimension of visual feature. Similarly, we use $\X^t\in\mathcal{R}^{d\times n^t}$ to denote the visual features of test images from $C^t$ test categories, in which $n^t$ is the number of test images.
Suppose that each category is associated with a $m$-dim semantic representation, the semantic representation matrix of test (\resp, auxiliary) categories is represented as $\bar{\A}^t\in\mathcal{R}^{m\times C^t}$ (\resp, $\bar{\A}^a\in\mathcal{R}^{m\times C^a}$). Then, the semantic representation matrix of well-labeled training data is represented as $\A^a\in\mathcal{R}^{m\times n^a}$, in which each column is the semantic representation of the category that each training instance belongs to. In a similar way, the semantic representation matrix of test data is given by $\A^t\in\mathcal{R}^{m\times n^t}$, which needs to be learnt. After obtaining $\A^t$, the category labels of test data can be inferred by comparing $\A^t$ with $\bar{\A}^t$. 

With the aim to transfer knowledge from auxiliary categories to test categories, inspired by~\cite{kodirov2015unsupervised}, we learn one visual-semantic dictionary $\D^a\in\mathcal{R}^{d\times m}$ (\resp, $\D^t\in\mathcal{R}^{d\times m}$) for auxiliary categories (\resp, test categories), expecting $\D^a$ and $\D^t$ to be close to each other using a co-regularizer $\|\D^t-\D^a\|_F^2$. The visual-semantic dictionary $\D^t$ (\resp, $\D^a$) maps from semantic representation space to visual feature space with the mapping error $\|\X^t-\D^t\A^t\|_F^2$ (\resp, $\|\X^a-\D^a\A^a\|_F^2$). It is worth mentioning that we adopt dictionary learning based method because it lays the foundation for unifying WSL and ZSL in a coherent manner.

The knowledge transfer has two steps. In the first step, the dictionary of auxiliary categories $\D^a$ is learnt as follows,
\begin{eqnarray} \label{eqn:first_stage}
\min_{\D^a} &&  \frac{1}{2}\|\X^a-\D^a\A^a\|_F^2 + \frac{1}{2} \|\D^a\|_F^2, 
\end{eqnarray}
\noindent in which $\|\X^a-\D^a\A^a\|_F^2$ is the mapping error of well-labeled training images from auxiliary categories and $\|\D^a\|_F^2$ is a penalty term controlling the complexity of $\D^a$.  

In the second step, we aim to learn the dictionary of test categories $\D^t$ and semantic representations of test instances $\A^t$. Specifically, we minimize the mapping error $\|\X^t\!-\!\D^t\A^t\|_F^2$ on the test images, similar to the first step. Besides, we enforce $\D^t$ to be close to $\D^a$ based on $\|\D^t\!-\!\D^a\|_F^2$, in which way the knowledge of auxiliary categories can be transferred to test categories. Moreover, considering that the semantic representations of the test instances belonging to the same category should be similar to each other, we expect $\A^t$ to be low-rank, which encourages the similar semantic representations of test instances to be grouped together. To achieve this goal, we bring in a nuclear norm~\cite{recht2010guaranteed} regularizer $\|\A^t\|_*$, which enforces $\A^t$ to be low-rank. To this end, the objective function of the second step is given by
\begin{eqnarray} \label{eqn:second_stage}
\min_{\D^t,\A^t}\frac{1}{2}\|\X^t\!-\!\D^t\A^t\|_F^2 \!+\! \frac{\lambda_1}{2}\|\D^t\!-\!\D^a\|_F^2 \!+\! \lambda_2\|\A^t\|_*,\label{eqn:WSZSL1_st}
\end{eqnarray}
\noindent in which $\lambda_1$ and $\lambda_2$ are trade-off parameters. 

\subsection{Exploiting Web Images from Test Categories} \label{sec:method_web_image}
Besides well-labeled training images from auxiliary categories, we additionally crawl web images by using $C^t$ test category names as queries to form the web training set. The visual features of web images are represented as $\X^w\in\mathcal{R}^{d\times n^w}$, in which $n^w$ is the number of web images. Analogous to $\A^a$, the semantic representation matrix of web images is given by $\A^w \in\mathcal{R}^{m\times n^w}$, in which each column is the semantic representation of the category that each web image is associated with (the category labels of web images may be inaccurate). Since web images and test images are from the same set of test categories, the same dictionary $\D^t$ is applied to the web images, leading to the mapping error of web images $\|\X^w-\D^t\A^w\|_F^2$. Recall that we need to address two issues when learning from web images: label noise and domain shift, which will be detailed next.

To account for the label noise of web images, we replace the mapping error, \ie, Frobenius norm regularizer $\|\X^w-\D^t\A^w\|_F^2$, with $L_{2,1}$ norm regularizer $\|\X^w-\D^t\A^w\|_{2,1}$. The $L_{2,1}$ norm of a matrix $\X$ is defined as $\sum_{i} \|\x_i\|_2$, in which $\x_i$ is each column in $\X$. $L_{2,1}$ norm encourages column-sparsity~\cite{yuan2006model}, that being said, some columns have exceeding zero entries. After employing $L_{2,1}$ norm, $\X^w-\D^t\A^w$ is expected to be column-sparse. The columns with exceeding non-zero (\resp, zero) entries correspond to the outliers (\resp, non-outliers), which is granted larger (\resp, smaller) tolerance of error. 
In this way, we suppress the label noise of web images and learn a more robust dictionary $\D^t$ on test categories. 

To account for the domain shift between web images (\ie, $\X^w$) and test images (\ie, $\X^t$), we employ an Maximum Mean Discrepancy (MMD)~\cite{Huang2007KMM} based regularizer $\|\frac{1}{n^w}\X^w\bm{\theta}\!-\!\frac{1}{n^t}\X^t\1\|^2$ with the weight vector $\bm{\theta}$ to be learnt. The idea of MMD-based regularizer is to reduce the distance between the center of weighted web images (\ie, $\frac{1}{n^w}\X^w\bm{\theta}$) and the center of test images (\ie, $\frac{1}{n^t}\X^t\1$). This is accomplished by assigning higher weights on the web images which are closer to the center of test images. In other words, we identify the web images which are more likely to be sampled from the data distribution of test images, by learning the weight vector $\bm{\theta}$.

To take full advantage of the weight vector $\bm{\theta}$, we expect to identify the web images with not only closer distribution to test images but also relatively accurate labels. Specifically, besides the MMD-based regularizer $\|\frac{1}{n^w}\X^w\bm{\theta}\!-\!\frac{1}{n^t}\X^t\1\|^2$, we also employ the weights $\bm{\theta}$ in the $L_{2,1}$ norm regularizer $\|(\X^w\!-\!\D^t\A^w)\bm{\Theta}\|_{2,1}$, in which we use a diagonal matrix $\bm{\Theta}$ with the diagonal being $\bm{\theta}$ for ease of representation. By minimizing $\|(\X^w\!-\!\D^t\A^w)\bm{\Theta}\|_{2,1}$, lower (\resp, higher) weights are prone to be assigned to the columns of $\X^w\!-\!\D^t\A^w$ with exceeding non-zero (\resp, zero) entries, which correspond to the outliers (\resp, non-outliers). In this way, we collaboratively account for label noise and domain shift with the importance weight vector $\bm{\theta}$ shared by two regularizers. In contrast, most existing works address these two issues separately. 

From another perspective, since the dictionary $\D^t$ used in $\|(\X^w\!\!-\!\D^t\A^w)\bm{\Theta}\|_{2,1}$ is enforced to be close to the dictionary of auxiliary categories $\D^a$, auxiliary categories actually assist in dealing with the label noise of web images.
To this end, we extend (\ref{eqn:second_stage}) by using web images as follows,
\begin{eqnarray} \label{eqn:WSZSL1}
\min_{\D^t,\A^t,\bm{\theta}}\!\!\!\!\!\!\!\!\! && \frac{1}{2}\|\X^t\!-\!\D^t\A^t\|_F^2 \!+\! \frac{\lambda_1}{2}\|\D^t\!-\!\D^a\|_F^2 \!+\! \lambda_2\|\A^t\|_*\nonumber\\
&&\!\!\!\!\!\!\!\!\!\!\!\!\!\!\!\!\!\!\!\!\!\!+ \frac{\lambda_3}{2} \|\frac{1}{n^w}\X^w\bm{\theta}\!-\!\frac{1}{n^t}\X^t\1\|^2 \!+\! \lambda_4\|(\X^w\!\!-\!\!\D^t\A^w)\bm{\Theta}\|_{2,1},\\
\mbox{s.t.} &&\1'\bm{\theta}=n^w,\quad \0\leq\bm{\theta}\leq b\1,\label{eqn:WSZSL1_st}
\end{eqnarray}
\noindent where $\lambda_3$, $\lambda_4$, and $b$ are newly introduced trade-off parameters. Note that we impose a sum constraint and a box constraint on $\bm{\theta}$ in (\ref{eqn:WSZSL1_st}), in which $b$ is the upper bound of importance weights. The problem in (\ref{eqn:WSZSL1}) is nontrivial to solve because of the $L_{2,1}$ norm and low-rank regularizer. Hence, a novel solution is developed based on inexact Augmented Lagrange Multiplier (ALM)~\cite{boyd2011distributed}. For ease of optimization, an intermediate variable $\Z^t$ (\resp, $\E^w$ ) is introduced to replace $\A^t$ in $\|\A^t\|_*$ (\resp, $(\X^w-\D^t\A^w)\bm{\Theta}$) in (\ref{eqn:WSZSL1}). At the same time, we enforce $\Z^t$ (\resp, $\E^w$ ) to be close to $\A^t$ (\resp, $(\X^w-\D^t\A^w)\bm{\Theta}$). Then, the problem in (\ref{eqn:WSZSL1}) can be rewritten as
\begin{eqnarray} \label{eqn:WSZSL2}
\min_{\D^t,\A^t,\bm{\theta}}\!\!\!\!\!\!\!\! && \frac{1}{2}\|\X^t\!-\!\D^t\A^t\|_F^2 \!+\! \frac{\lambda_1}{2}\|\D^t\!-\!\D^a\|_F^2 \!+\! \lambda_2\|\Z^t\|_* \nonumber\\
&&\!+ \frac{\lambda_3}{2} \|\frac{1}{n^w}\X^w\bm{\theta}\!-\!\frac{1}{n^t}\X^t\1\|^2 \!+\!\lambda_4\|\E^w\|_{2,1},\\
\mbox{s.t.} &&\1'\bm{\theta}=n^w,\quad\0\leq\bm{\theta}\leq b\1,\nonumber\\
&& \E^w=(\X^w-\D^t\A^w)\bm{\Theta},\label{eqn:st_Ew}\\
&& \Z^t = \A^t.\label{eqn:st_Zt}
\end{eqnarray}

\noindent Then, after introducing the Lagrangian multiplier $\R$ (\resp, $\T$) for the constraint in (\ref{eqn:st_Ew}) (\resp, (\ref{eqn:st_Zt})), we tend to minimize the augmented Lagrangian form of (\ref{eqn:WSZSL2}):
\begin{eqnarray} \label{eqn:WSZSL2_lag}
\mathcal{L}_{\stackrel{\D^t,\A^t,\Z^t}{\E^w, \bm{\theta}\in\bm{\mathcal{S}}}} =\!\!\!\! &&\!\!\!\!\!\!\!\!\frac{1}{2}\|\X^t\!-\!\D^t\A^t\|_F^2 \!+\! \frac{\lambda_1}{2}\|\D^t\!-\!\D^a\|_F^2 \!+\!\lambda_2\|\Z^t\|_*\nonumber\\
&&\!\!\!\!\!\!\!\!\!\!\!\!\!\!\!\!\!\!\!\!\!\!\!\!\!\!\!\!\!\!\!\!\!\!\!\!\!\!+ \frac{\lambda_3}{2} \|\frac{1}{n^w}\X^w\bm{\theta}-\frac{1}{n^t}\X^t\1\|^2+ \lambda_4\|\E^w\|_{2,1}\nonumber\\
&&\!\!\!\!\!\!\!\!\!\!\!\!\!\!\!\!\!\!\!\!\!\!\!\!\!\!\!\!\!\!\!\!\!\!\!\!\!\!+\frac{\mu}{2}\|\E^w\!\!-\!(\X^w\!\!-\!\D^t\A^w)\bm{\Theta}\|_F^2\!+\!\left\langle\R,\!\E^w\!\!-\!(\X^w\!\!-\!\D^t\A^w)\bm{\Theta} \right\rangle\nonumber\\
&&\!\!\!\!\!\!\!\!\!\!\!\!\!\!\!\!\!\!\!\!\!\!\!\!\!\!\!\!\!\!\!\!\!\!\!\!\!\!+\frac{\mu}{2}\|\A^t-\Z^t\|_F^2+\left\langle\T, \A^t-\Z^t\right\rangle, 
\end{eqnarray}
\noindent in which $\bm{\mathcal{S}}=\{\bm{\theta}|1'\bm{\theta}=n^w, \0\leq\bm{\theta}\leq b\1 \}$ is the feasible set of $\bm{\theta}$, and $\mu$ is a penalty parameter. We update the variables $\{\E^w,\Z^t,\D^t,\A^t,\bm{\theta}\}$, the Lagrangian multipliers $\{\R,\T\}$, and the penalty parameter $\eta$ iteratively until the termination criterion is satisfied. The technical details of updating these variables are left to Appendix~\ref{sec:appendix_WSZSL}. By minimizing (\ref{eqn:WSZSL2_lag}), we can acquire the semantic representations of test instances $\A^t$. 

With the semantic representations of test images $\A^t$ and test categories $\bar{\A}^t$, we use nearest neighbor (NN) classifier for final category prediction, following the strategy in~\cite{kodirov2015unsupervised}. In particular, we compare the semantic representation of each test instance (\ie, each column in $\A^t$) with that of each test category (\ie, each column in $\bar{\A}^t$), and each test instance is assigned to the nearest test category.

\subsection{Extension with Privileged Information} \label{sec:method_PI}
As mentioned in Section~\ref{sec:intro}, web images are usually associated with additional textual information, which is unavailable for test images. The additional information which is only available for the training data but not available for the test data is referred to as privileged information. Learning using privileged information was initially proposed in~\cite{LUPIparadigm}. In particular, they extend SVM to SVM+ by using privileged information, in which the slack variable $\xi_i$ in SVM is replaced by a slack function $\xi(\tx_i)$ based on the privileged information $\tx_i$. Formally, the objective function of SVM+ is
\begin{eqnarray} \label{eqn:SVM+}
\min_{\w,b,\tilde{\w},\tilde{b}} &&\frac{1}{2}(\|\w\|^2+\|\tilde{\w}\|^2) + \sum_{i=1}^n \xi(\tx_i)\\
\mbox{s.t.} && y_i(\w\x_i+b)\geq 1-\xi(\tx_i),\quad\forall i,\nonumber\\
&&\xi(\tx_i)\geq 0,\quad\forall i,\nonumber\\
&&\xi(\tx_i)=\tw'\tx_i+\tilde{b},\quad\forall i.
\end{eqnarray}
where $\w$ (\resp, $\tw$) and $b$ (\resp, $\tilde{b}$) are the weight vector and bias of classification (\resp, slack) function based on primal feature $\x_i$ (\resp, privileged information $\tx_i$), and $y_i$ is the binary label. Recall that in conventional SVM, the slack variable $\xi_i$ is used to model the difficulty of classifying each training instance. In analogy to $\xi_i$, SVM+ relies on the slack function $\xi(\tx_i)$ to model the classification difficulty, in which privileged information $\tx_i$ plays a role of teacher in the training process. To be exact, if one training instance is difficulty to classify, the value of its slack function $\xi(\tx_i)$ is allowed to be very large for tolerance of error. Otherwise, the value of its slack function $\xi(\tx_i)$ is enforced to be very small. Note that $\tw$ and $\tilde{b}$ are automatically learnt by solving (\ref{eqn:SVM+}).

Inspired by SVM+, we also use the slack function based on privileged information to model the mapping difficulty. Recall that in our problem, we learn a mapping $\D^t$ on web images to map from semantic space $\A^w$ to visual space $\X^w$, leading to the mapping error $\X^w-\D^t\A^w$, which stands for the mapping difficulty of each web image. Similar to SVM+, we tend to use privileged information (\ie, textual information) to approximate the mapping error. After denoting the aggregated textual feature of web images as $\tilde{\X}^w \in \mathcal{R}^{\tilde{d}\times n^w}$ with $\tilde{d}$ being the dimensionality of textual feature, we learn a slack function based on $\tilde{\X}^w$ to estimate the mapping error. Simply, we learn a matrix $\tilde{\W}\in\mathcal{R}^{d\times \tilde{d}}$ and enforce $\tilde{\W}\tilde{\X}^w$ to be close to the mapping error $\X^w-\D^t\A^w$. On one hand, the slack function $\tilde{\W}\tilde{\X}^w$ allows the web training images with great mapping difficulty to have large mapping error. On the other hand, when the number of training instances is not very large, the slack function can help avoid over-fitting by regulating the mapping error~\cite{YangZCO17}.

To this end, we actually control the mapping error in two ways: 1) assign different weights $\bm{\Theta}$ on the mapping error of different web images; 2) use additional textual information to approximate the mapping error of each web image.
After adding a new regularizer $\|(\X^w-\D^t\A^w)-\tilde{\W}\tilde{\X}^w\|_F^2$ to (\ref{eqn:WSZSL2}), our new objective function can be written as
\begin{eqnarray} \label{eqn:WSZSL2_PI}
\min_{\stackrel{\D^t,\A^t,}{\bm{\theta},\tilde{\W}}}\!\!\!\!\!\!\!\! && \frac{1}{2}\|\X^t\!-\!\D^t\A^t\|_F^2 \!+\! \frac{\lambda_1}{2}\|\D^t\!-\!\D^a\|_F^2 \!+\! \lambda_2\|\Z^t\|_* \nonumber\\
&&\!+ \frac{\lambda_3}{2} \|\frac{1}{n^w}\X^w\bm{\theta}\!-\!\frac{1}{n^t}\X^t\1\|^2 \!+\!\lambda_4\|\E^w\|_{2,1}\nonumber\\
&& + \frac{\gamma}{2}\|(\X^w-\D^t\A^w)-\tilde{\W}\tilde{\X}^w\|_F^2,\\
\mbox{s.t.} &&\1'\bm{\theta}=n^w,\quad\0\leq\bm{\theta}\leq b\1,\nonumber\\
&& \E^w=(\X^w-\D^t\A^w)\bm{\Theta},\quad \Z^t = \A^t,\nonumber
\end{eqnarray}
in which $\gamma$ is a trade-off parameter. Compared with (\ref{eqn:WSZSL2}), we have one more variable $\tilde{\W}$ to learn. The problem in (\ref{eqn:WSZSL2_PI}) can be solved similarly to (\ref{eqn:WSZSL2}) except updating $\D^t$ and $\tilde{\W}$. We leave the technical details of solving (\ref{eqn:WSZSL2_PI}) to Appendix~\ref{sec:appendix_WSZSL_PI}. After solving (\ref{eqn:WSZSL2_PI}), we obtain the semantic representations of test instances $\A^t$ and the testing procedure is the same as in Section~\ref{sec:method_web_image}. Note that we do not need textual information of test instances in the testing stage.

\section{Experiments}
In this section, we evaluate our methods for fine-grained image classification with or without privileged information on three benchmark datasets. Besides, we conduct extra experiments under the generalized setting in which the test instances may come from both auxiliary categories and test categories. Moreover, we also provide adequate ablation study and qualitative analysis.

\subsection{Fine-grained Image Classification} \label{sec:exp_wo_PI}
\noindent\textbf{Datasets:} Experiments are conducted on the following three datasets which are popular in the zero-shot learning (ZSL) community, since our learning scenario can be treated as ZSL with additional web training images for unseen categories, as mentioned in Section~\ref{sec:intro}.

\noindent 1) CUB~\cite{WahCUB_200_2011}: Caltech-UCSD Bird (CUB) consists of $11,788$ images from $200$ bird species. Following \cite{akata2013label}, we adopt the standard train-test split with $150$ auxiliary (\resp, $50$ test) categories.

\noindent 2) SUN~\cite{xiao2010sun}: In Scene UNderstanding (SUN) attribute dataset, each scene category has 20 images. Following \cite{jayaraman2014zero}, we adopt the standard train-test split with $707$ auxiliary (\resp, $10$ test) categories.

\noindent 3) Dogs~\cite{KhoslaYaoJayadevaprakashFeiFei_FGVC2011}: Stanford Dogs dataset is composed of $19,501$ images distributed in $113$ dog breeds. We use the train-test split provided in \cite{akata2015evaluation}, \ie, $85$ auxiliary (\resp, $28$ test) categories.

\noindent 4) Flickr image dataset: We construct the web training set by ourselves. Particularly, for each benchmark dataset (\ie, CUB, SUN, and Dogs), we use the names of test categories as queries to collect the top ranked $100$ images from Flickr website for each category after performing PCA-based near-duplicate removal~\cite{zhou2016places}.

\noindent\textbf{Visual Features and Semantic Representations}: We extract visual features for all images and semantic representations for all categories.

\noindent 1) Visual features: For each image, we use $4,096$-dim output of the $6$-th layer of VGG~\cite{simonyan2014very} model pretrained on ImageNet dataset as its visual feature.

\noindent 2) Semantic representations: We employ two types of word vectors: GloVe~\cite{pennington2014glove} and Word2Vec~\cite{mikolov2013distributed}, in which each word is associated with a real-valued vector.
We train Word2Vec and GloVe linguistic models based on the latest Wikipedia corpus, with the dimension of word vector set as $400$. Then, for each category, two word vectors corresponding to the category name from Word2Vec and GloVe models are concatenated as the category-level semantic representation, yielding an $800$-dim vector. When one category has the name with more than one word, we simply use the average of the semantic representations corresponding to all words as its final semantic representation.

\begin{table}[t]
\caption{Accuracies (\%) of different methods on three datasets. The best results are highlighted in boldface.}
\setlength{\tabcolsep}{5pt}
\label{tab:exp_results}
\centering
\begin{tabular}{|c|c|c|c|c|}
\hline
Dataset & CUB & SUN & Dogs & Avg \\
\hline
LR  & 68.39 & 62.50 & 77.67 & 69.52\\
\hline
KMM~\cite{Huang2007KMM} & 70.54 & 64.00 & 79.16 & 71.23\\
GFK~\cite{Gong2012gfk} & 70.37 & 62.50 & 79.51 & 70.79\\
SA~\cite{SA} & 68.67 & 63.00 & 80.18 & 70.62\\
TCA~\cite{Pan2011TCA}  & 68.56 & 63.00 & 80.22 & 70.59\\
CORAL~\cite{sun2016return} & 69.04 & 63.50 & 80.37 & 70.97 \\
\hline
NEIL~\cite{chen2013neil} & 69.08 & 63.00 & 80.16 & 70.74\\
Bergamo and Torresani~\cite{bergamo2010exploiting}  & 70.13 & 64.00 & 78.64 & 70.93\\
WSDG~\cite{niu2015visual}   & 70.61 & 66.00 & 80.20 & 72.27\\
Sukhbaatar~\etal~\cite{sukhbaatar2014training}  & 70.47 & 64.50 & 81.15 & 72.04\\
Xiao~\etal~\cite{xiao2015learning} & 70.92 & 65.50 & 81.67 & 72.69 \\
\hline
ESZSL~\cite{romera2015embarrassingly} & 38.08 & 65.00 & 37.21 & 46.77\\
LatEm~\cite{xian2016latent}  & 35.15 & 66.50 & 35.99 & 45.88\\
SJE~\cite{akata2015evaluation}  & 42.65 & 71.50 & 34.85 & 49.67\\
DAP/IAP~\cite{lampert2014attribute}  & 28.91 & 57.50 & 33.15 & 39.85\\
Changpinyo~\etal~\cite{changpinyo2016synthesized}  & 41.83 & 72.00 & 39.91 & 51.25\\
Li~\etal~\cite{li2015semi} & 32.36 & 72.50 & 43.15 & 49.34\\
Kodirov~\etal~\cite{kodirov2015unsupervised} & 47.53 & 71.00 & 47.32 & 55.28\\
Zhang and Saligrama~\cite{zhang2016zero} & 44.08 & 76.50 & 48.09 & 56.23\\
Xu~\etal~\cite{xu2017transductive}  & 45.72 & 71.50 & 39.85 & 52.36\\
Shojaee and Baghshah~\cite{shojaee2016semi} & 46.68 & 71.00 & 48.82 & 55.50\\
Zhang and Koniusz~\cite{zhang2018zero} & 47.66 & 73.00 & 43.98 & 54.88 \\
SE-GZSL~\cite{kumar2018generalized} & 46.84 & 75.00 & 43.91 & 55.25\\
\hline
WSL+ZSL & 72.21 & 78.50 & 81.90 & 77.53 \\
\hline
Ours\_WSL   & 69.42 & 65.50 & 80.43 & 71.78\\
Ours\_ZSL   & 47.94 & 71.50 & 47.70 & 55.71\\
Ours\_sim1   & 72.72 & 83.50 & 85.04 & 80.42\\
Ours\_sim2   & 76.00 & 79.50 & 83.75 & 79.75\\
Ours  & \textbf{76.47} & \textbf{84.50} & \textbf{85.16} & \textbf{82.04}\\
\hline
\end{tabular}
\end{table}

\noindent\textbf{Baselines:}  We compare our approach with three sets of baselines: WSL baselines, ZSL baselines, and domain adaptation (DA) baselines. To the best of our knowledge, no existing method can jointly utilize web data and auxiliary categories, so we combine the most competitive ZSL and DA/WSL baselines by simply averaging their test decision values as the combo baseline. Intuitively, the combo baseline should be the strongest baseline because it utilizes both web images and auxiliary categories.

For WSL baselines, we compare with NEIL~\cite{chen2013neil}, Bergamo and Torresani~\cite{bergamo2010exploiting}, WSDG~\cite{niu2015visual}, sukhbaatar~\etal~\cite{sukhbaatar2014training}, and Xiao~\etal~\cite{xiao2015learning}. Note that Xiao~\etal~\cite{xiao2015learning} leverages manually cleaned web data when training network and computing confusion matrix, which is not available in our application. Therefore, for fair comparison, we run \cite{xiao2015learning} without using manually cleaned web data in the training process and estimate the confusion matrix based on semantic representations.

For ZSL baselines, we compare with the standard ZSL methods ESZSL~\cite{romera2015embarrassingly}, LatEm~\cite{xian2016latent},
SJE~\cite{akata2015evaluation}, DAP/IAP~\cite{lampert2014attribute}, Changpinyo~\etal~\cite{changpinyo2016synthesized}, Zhang and Koniusz~\cite{zhang2018zero} as well as transductive/semi-supervised ZSL methods Li~\etal~\cite{li2015semi}, Kodirov~\etal~\cite{kodirov2015unsupervised}, Zhang and Saligrama~\cite{zhang2016zero}, Xu~\etal~\cite{xu2017transductive}, Shojaee and Baghshah~\cite{shojaee2016semi}, SE-GZSL~\cite{kumar2018generalized} as baselines. The difference between transductive/semi-supervised ZSL approaches and standard ZSL approaches lies in whether unlabeled test data are available in the training phase.

For DA baselines, we compare with KMM~\cite{Huang2007KMM}, GFK~\cite{Gong2012gfk}, SA~\cite{SA}, TCA~\cite{Pan2011TCA}, and CORAL~\cite{sun2016return}, in which web training images and test images are regarded as the source domain and the target domain respectively.

For the combo baseline, we select the most competitive WSL baseline \cite{xiao2015learning} and ZSL baseline \cite{zhang2016zero} based on their mean performance on three datasets, and average their test decision values, which is referred to as WSL+ZSL in Table~\ref{tab:exp_results}.

We also compare with one basic baseline LR, which simply learns a linear regressor based on web training images. Besides, to validate the WSL and ZSL components in our method (\ref{eqn:WSZSL1}), we report the results of our two special cases. Particularly, we remove the regularizer related to knowledge transfer (\ie, $\|\D^t-\D^a\|_F^2$) by setting $\lambda_1$ as $0$, and refer to this special case as Ours\_WSL. Similarly, we remove the regularizers using web data (\ie, $\|\frac{1}{n^w}\X^w\bm{\theta}\!-\!\frac{1}{n^t}\X^t\1\|^2$ and $\|(\X^w\!\!-\!\!\D^t\A^w)\bm{\Theta}\|_{2,1}$) by setting $\lambda_3$ and $\lambda_4$ as $0$, and this special case is referred to as Ours\_ZSL. Moreover, to validate some regularizers individually in our method (\ref{eqn:WSZSL1}), we further compare with our two simplified versions. Specifically, we remove the regularizer $\|\A^t\|_*$ (\resp, $\|\frac{1}{n^w}\X^w\bm{\theta}\!-\!\frac{1}{n^t}\X^t\1\|^2$) in (\ref{eqn:WSZSL1}) by setting $\lambda_2$ (\resp, $\lambda_3$) as $0$ and refer to this simplified version as Ours\_sim1 (\resp, Ours\_sim2). For all methods, we use multi-class accuracy as the evaluation metric. 

\noindent\textbf{Parameters:} Our method has trade-off parameters $b$, $\lambda_1$, $\lambda_2$, $\lambda_3$, and $\lambda_4$ in (\ref{eqn:WSZSL1}), which are determined by using the cross-validation strategy.
In particular, following \cite{shojaee2016semi}, we select the first $C^c$ categories according to default category indices from $C^a$ auxiliary categories as the validation categories, with $C^c$ satisfying $\frac{C^c}{C^a}=\frac{C^t}{C^a+C^t}$. It is worth mentioning that we need to additionally crawl web images for validation categories in order to use the cross-validation strategy. In the validation stage, we use $C^c$ categories as test categories and $C^a-C^c$ categories as auxiliary categories. Then, the optimal trade-off parameters are determined according to the validation performance through random search~\cite{bergstra2012random} within certain range. To be exact,
we empirically traverse the parameters $\lambda_1$, $\lambda_2$, $\lambda_3$, and $\lambda_4$ within the range $[10^{-3},10^{-2},\ldots,10^{3}]$, and traverse the parameter $b$ within the range $[1.5, 2.0, \ldots, 5.0]$. The range of $b$ is explained as follows. The upper bound of importance weights should be larger than one yet not too large based on the mild assumption that no web training image is far more important than others.

\begin{figure*}[t]
 	\centering
    \includegraphics[width=0.245\textwidth]{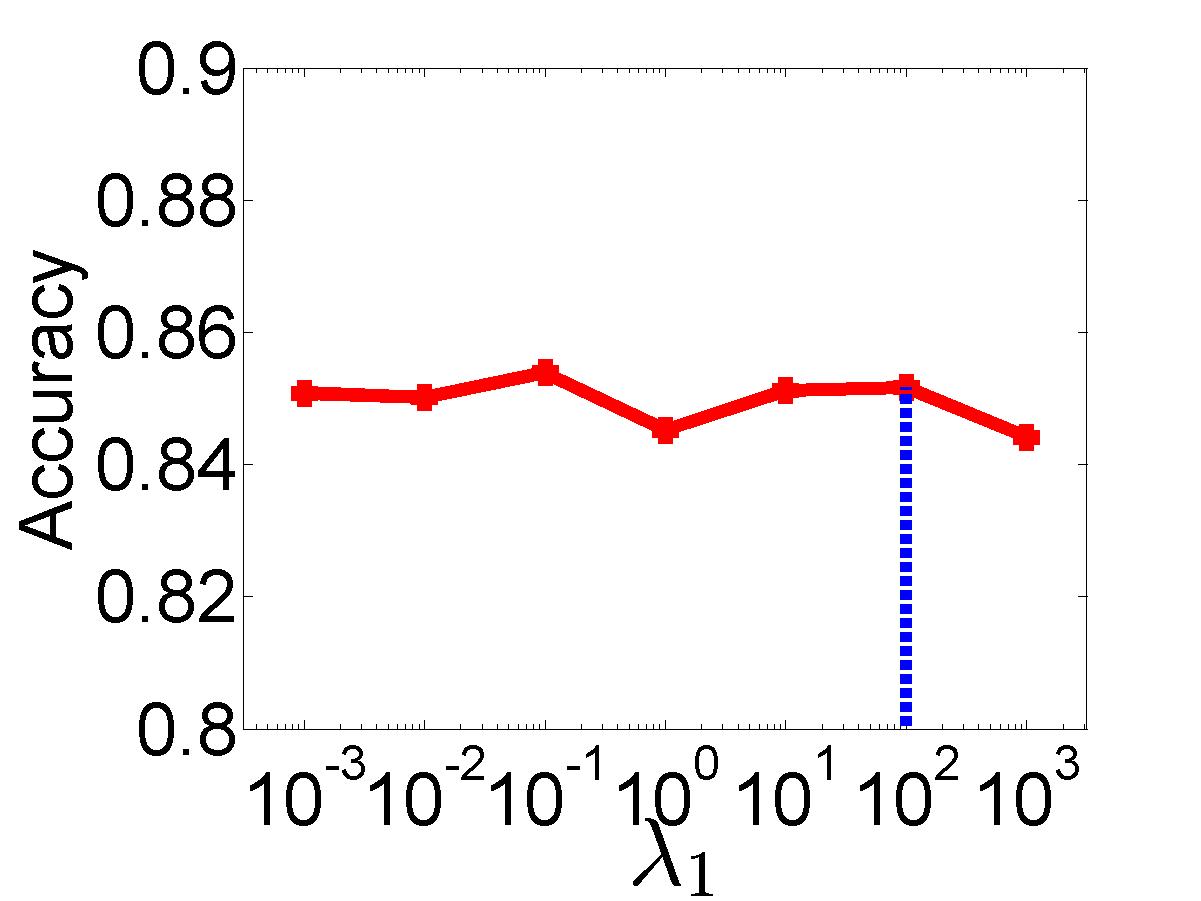}
	\includegraphics[width=0.245\textwidth]{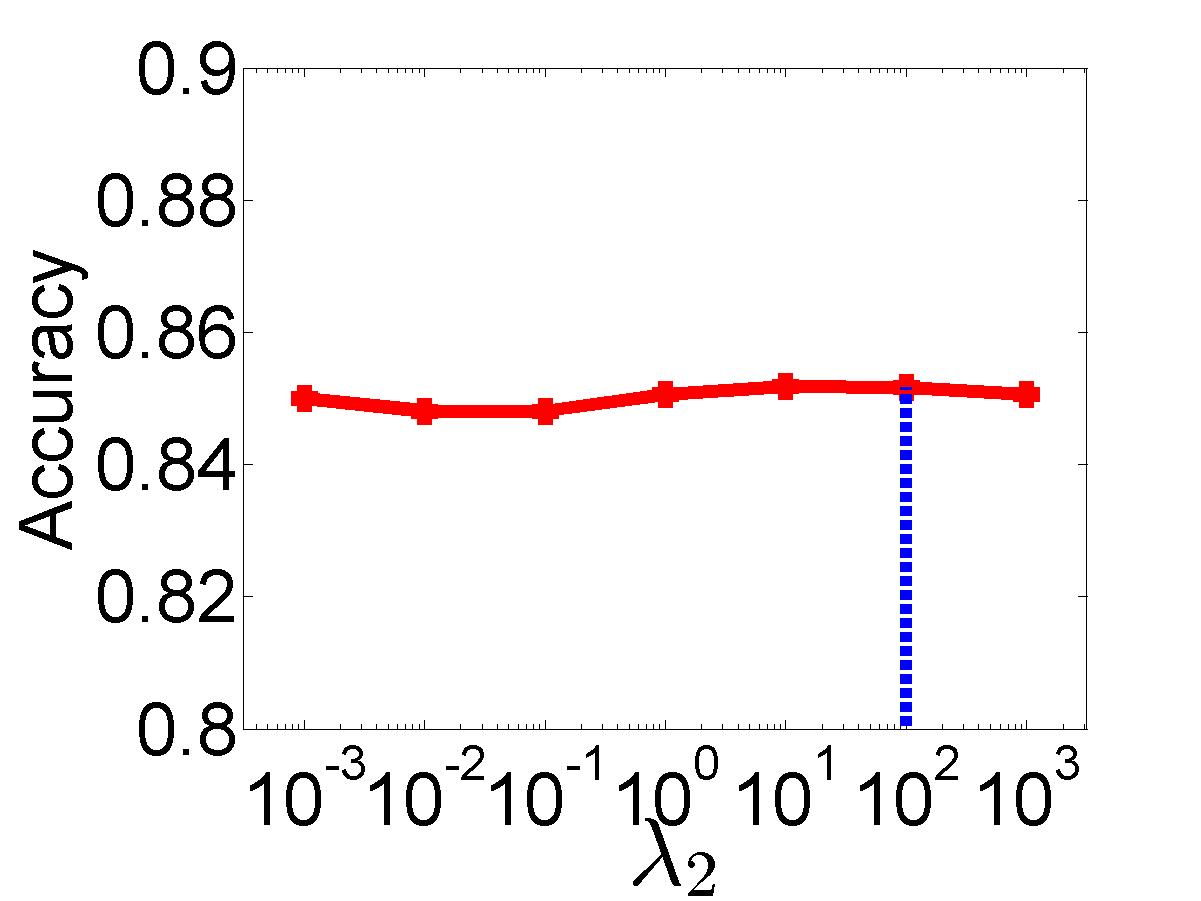}
	\includegraphics[width=0.245\textwidth]{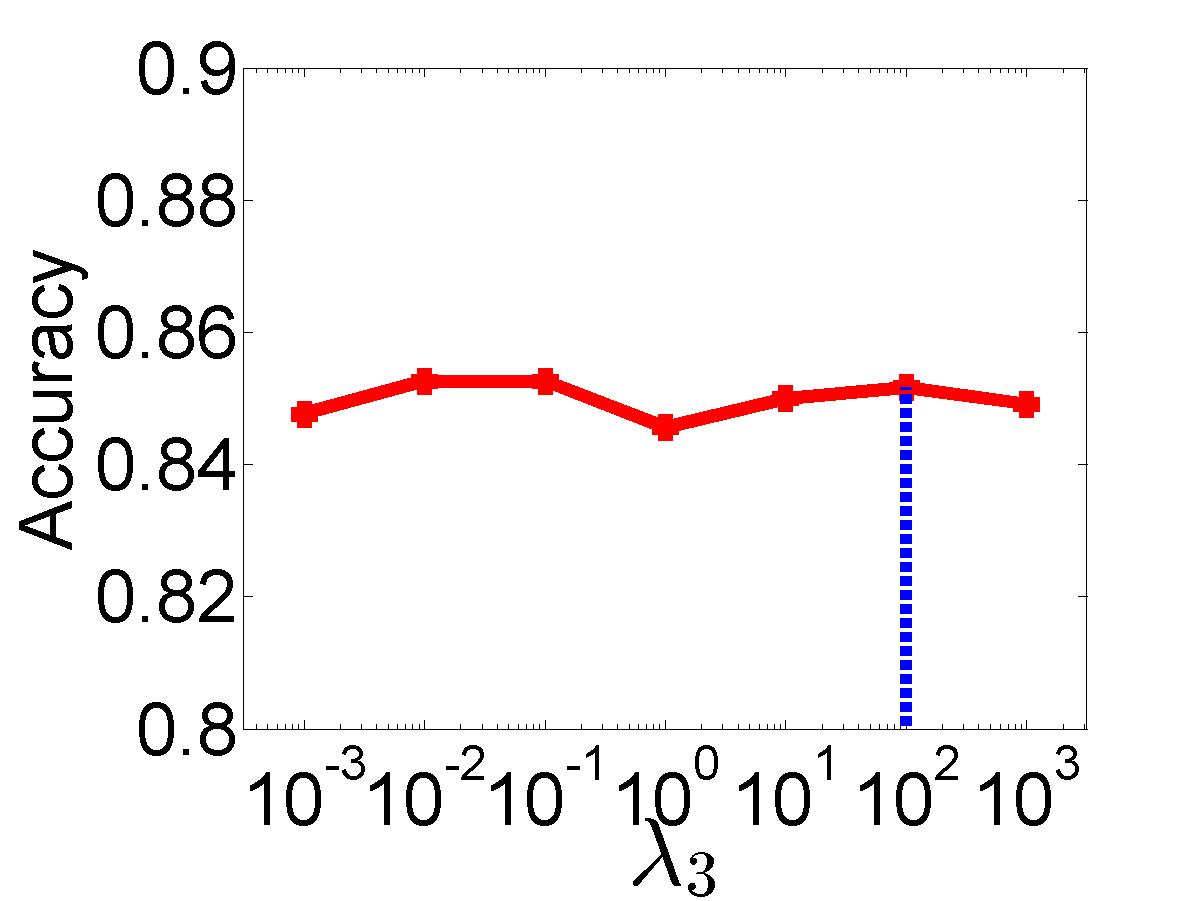}
	\includegraphics[width=0.245\textwidth]{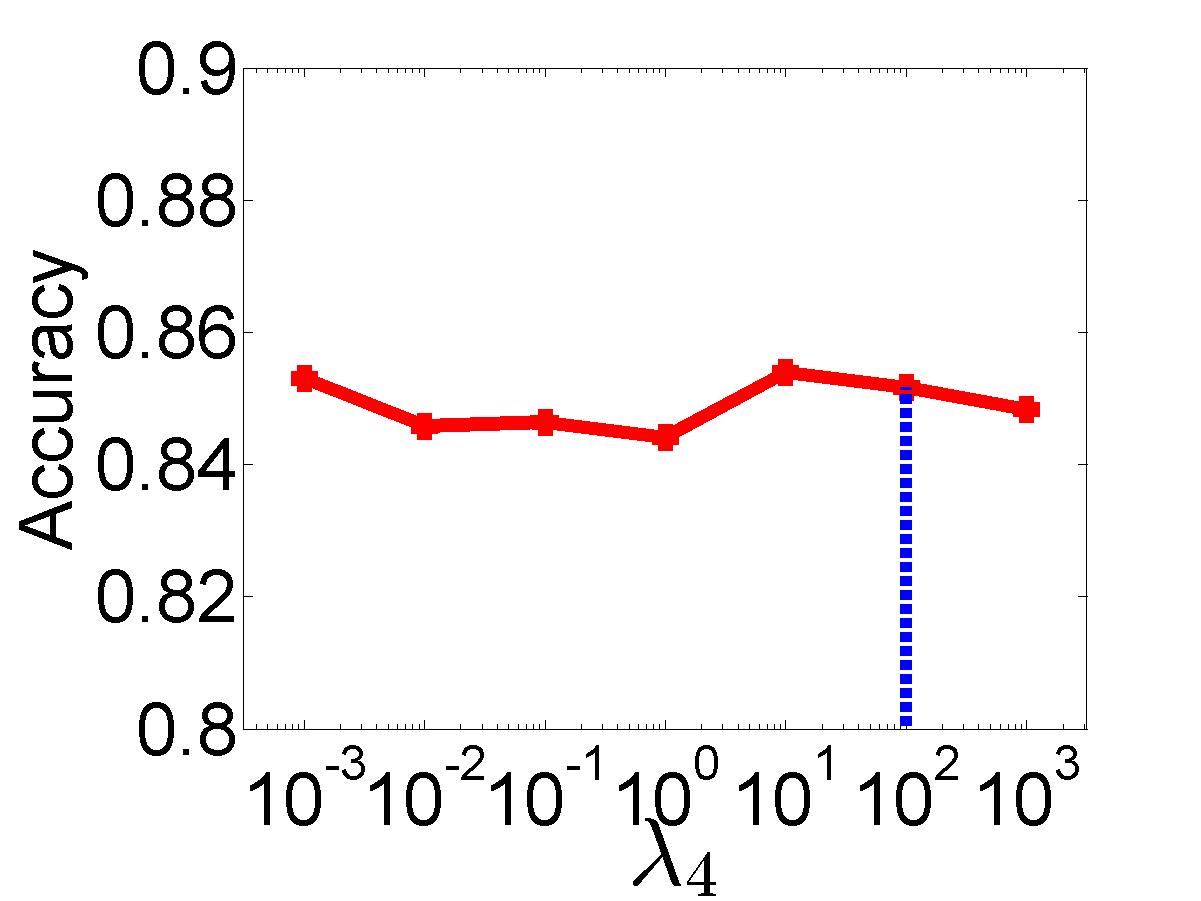}
    \caption{The performance variation of our method on the Dogs dataset by tuning one trade-off parameter and fixing other trade-off parameters as their optimal values. Vertical dashed lines indicate the optimal value of each trade-off parameter.}
    \label{fig:trade_off_parameters}
\end{figure*}

In fact, our method is relatively robust when setting the trade-off parameters within certain range. By taking the Dogs dataset as an example, we explore the performance variation of our approach \wrt one parameter while the other parameters remain fixed as their optimal values. It can be seen from Fig.~\ref{fig:trade_off_parameters} that our method is relatively robust when varying one parameter within the range $[10^{-3},10^{-2},\ldots,10^{3}]$ while fixing the other parameters as their optimal values.

\noindent\textbf{Experimental Results:} The experimental results of all methods are listed in Table~\ref{tab:exp_results}, based on which we have the following observations:

\noindent 1) The WSL and DA baselines are better than LR, which shows the advantage of coping with label noise or  domain shift. The ZSL baselines are worse than DA/WSL baselines on CUB and Dogs datasets, but generally better on the SUN dataset. There is no consistent winner between ZSL baselines and DA/WSL baselines because their performance highly depends on the purity of web images as well as the relation between auxiliary categories and test categories.

\noindent 2) The transductive/semi-supervised ZSL methods \cite{li2015semi,kodirov2015unsupervised,zhang2016zero,xu2017transductive,shojaee2016semi,kumar2018generalized} generally perform more favorably than the standard ZSL methods \cite{romera2015embarrassingly,xian2016latent,akata2015evaluation,lampert2014attribute,changpinyo2016synthesized,
zhang2018zero}, which demonstrates that it is useful to include unlabeled test data in the training phase for ZSL.

\noindent 3) Our method outperforms Ours\_WSL and Ours\_ZSL, which indicates the benefit of unifying WSL and ZSL. Our method also achieves better results than Ours\_sim1 and Ours\_sim2, which proves the effectiveness of our low-rank and MMD-based regularizers.

\noindent 4) It is worth mentioning that the focus of this paper is a new learning scenario for fine-grained classification by using both web data and auxiliary categories, instead of a state-of-the-art WSL or ZSL approach. So there is no guarantee that Ours\_WSL (\resp, Ours\_ZSL) can perform better than all WSL (\resp, ZSL) baselines. However, when utilizing both web data and auxiliary categories, our method achieves significant improvement over the combo baseline WSL+ZSL, which shows that a naive combination can hardly take full advantage of both web data and auxiliary categories. In contrast, we unify ZSL and WSL coherently in our method, which significantly advances fine-grained image classification.

\noindent\textbf{Computational Efficiency: }
By taking the CUB dataset as an example, we compare the running time of two special cases of our method (\emph{i.e.}, Ours\_WSL and Ours\_ZSL from Table 1) and their naive Combo (average $\mathbf{A}^t$ for prediction). The running time of Combo is the sum of running time of two special cases. We run all methods on the same server with Intel Xeon 3.33-GHz CPUs and 32-GB RAM in a single thread. The running time and accuracies of various approaches are reported in Table~\ref{tab:time_acc}, from which we can observe that our method is more effective and efficient than Combo.

\begin{table}[h]
\caption{Running time (s) and accuracies (\%) of different methods on the CUB dataset.}
\setlength{\tabcolsep}{4pt}
\label{tab:time_acc}
\centering
\small
\begin{tabular}{|c|c|c|c|c|}
\hline
Method & Ours\_WSL & Ours\_ZSL & Combo & Ours\\
\hline 
Time (s) & 1630.35 & 819.26 & 2449.61 & 1953.92 \\
\hline
Accuracies (\%)   & 69.42 & 47.94 & 70.58 & 76.47\\
\hline
\end{tabular}
\end{table}

\begin{figure}[t]
	\centering
    \includegraphics[width=0.45\textwidth]{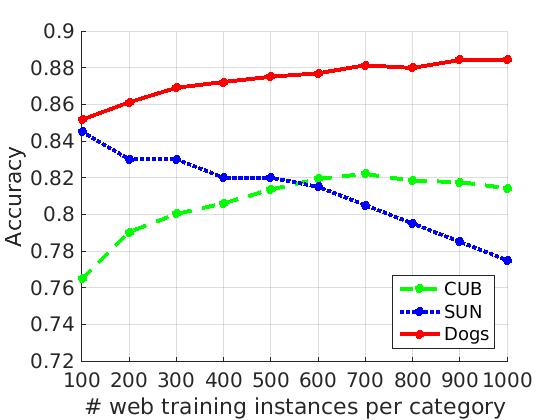}		
    \caption{The performance variation of our method \wrt different numbers of web training images per category.}
    \label{fig:web_training_num}
\end{figure}

\noindent\textbf{Utilizing More Web Images}: Since we only use $100$ web training images for each test category, it is interesting to explore whether the performance will keep increasing by using more web training images. We study the performance variation \wrt different numbers of web training images. Specifically, we crawl various numbers of web images for each test category
(\ie, $[100, 200, \ldots, 1000]$) to construct the web training set while keeping the other experimental settings unchanged. The accuracies with various
numbers of web training images on three datasets are plotted in Fig.~\ref{fig:web_training_num}, from which it can be seen that for the CUB and Dogs datasets, the accuracy increases as the number of  web training images grows within certain range. On the contrary, for the SUN dataset, the accuracy drops dramatically as the number of web training images grows. One possible explanation is that scene category names are more ambiguous than the dog/bird names. Furthermore, the scene category names in the SUN datasets are accompanied by additional ``in\_door" or ``out\_door" label, rendering it even more difficult to crawl semantically correct web images. 

\begin{figure}[t]
 	\centering
 	\begin{subfigure}[b]{0.0843\textwidth}
    		\includegraphics[width=1\textwidth]{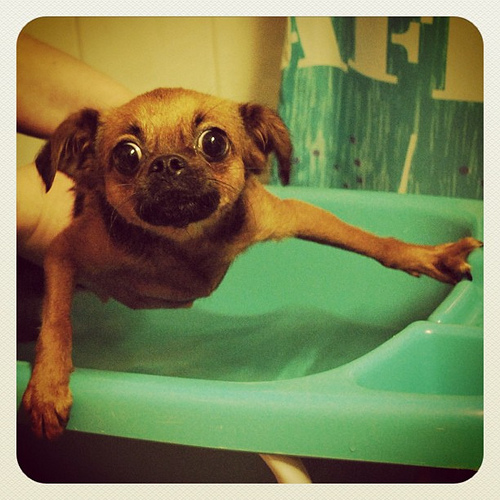}
		\caption{$1.46$}
    \end{subfigure}
    \begin{subfigure}[b]{0.0632\textwidth}
    		\includegraphics[width=1\textwidth]{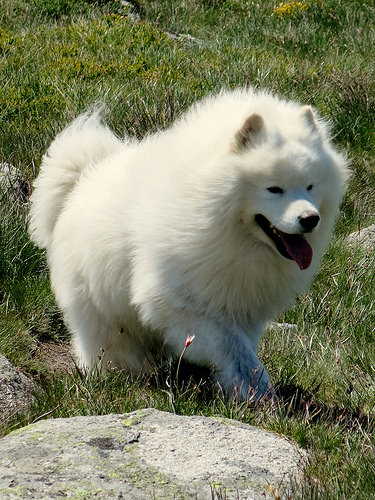}
		\caption{$1.46$}
    \end{subfigure}
    \begin{subfigure}[b]{0.1124\textwidth}
    		\includegraphics[width=1\textwidth]{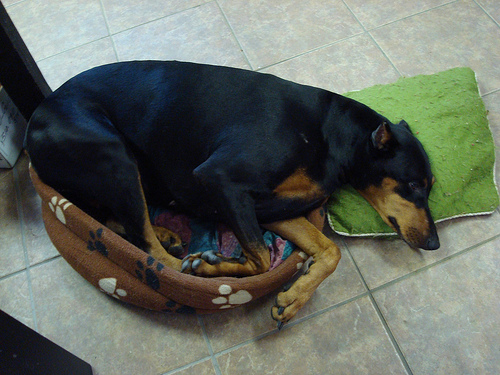}
		\caption{$1.35$}
    \end{subfigure}
    \begin{subfigure}[b]{0.1269\textwidth}
    		\includegraphics[width=1\textwidth]{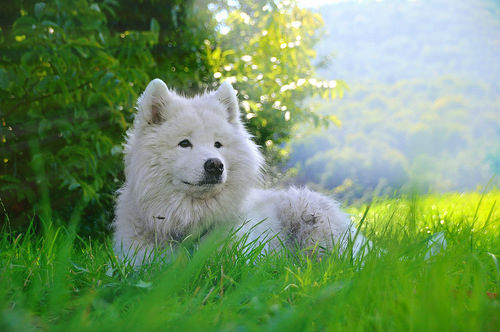}
		\caption{$1.35$}
    \end{subfigure}
    \begin{subfigure}[b]{0.0632\textwidth}
    		\includegraphics[width=1\textwidth]{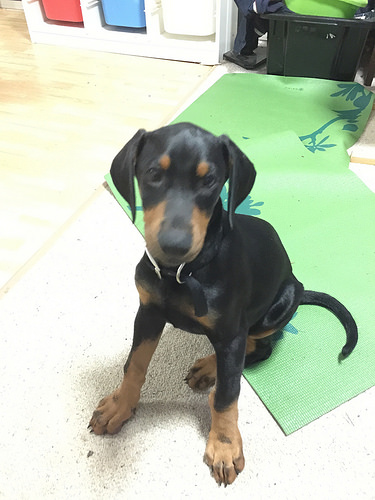}
		\caption{$1.34$}
    \end{subfigure}
    
    \begin{subfigure}[b]{0.0554\textwidth}
    		\includegraphics[width=1\textwidth]{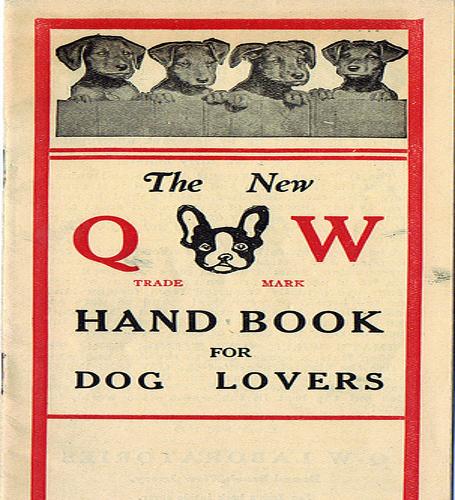}
		\caption{$0.75$}
    \end{subfigure}
    \begin{subfigure}[b]{0.0812\textwidth}
    		\includegraphics[width=1\textwidth]{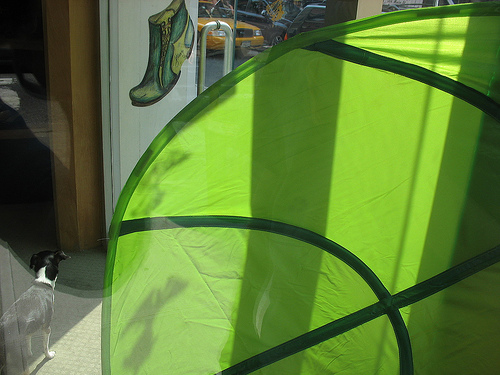}
		\caption{$0.75$}
    \end{subfigure}
    \begin{subfigure}[b]{0.0812\textwidth}
    		\includegraphics[width=1\textwidth]{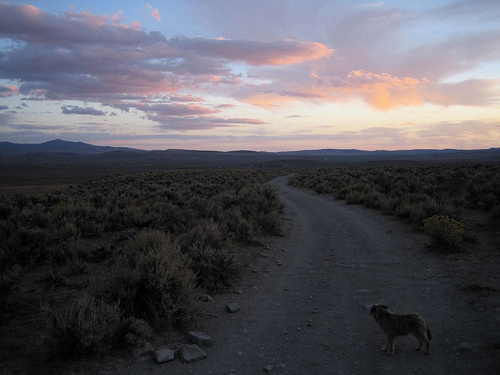}
		\caption{$0.75$}
    \end{subfigure}
    \begin{subfigure}[b]{0.1084\textwidth}
    		\includegraphics[width=1\textwidth]{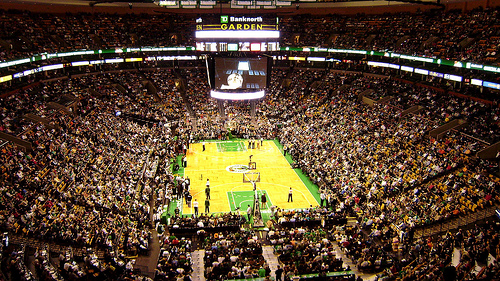}
		\caption{$0.74$}
    \end{subfigure}
    \begin{subfigure}[b]{0.1238\textwidth}
    		\includegraphics[width=1\textwidth]{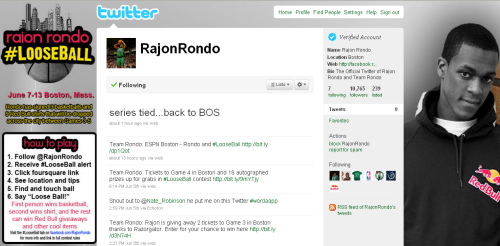}
		\caption{$0.72$}
    \end{subfigure}
    \caption{The web images in the top (\resp, bottom) row are associated with $5$ highest (\resp, lowest) weights based on the weight vector $\bm{\theta}$ learnt by our method.}
    \label{fig:weights}
\end{figure}

\noindent\textbf{Qualitative Analysis of Learnt Weights $\bm{\theta}$}:  In our method (\ref{eqn:WSZSL1}), higher weights are assumed to be assigned to the web training images with closer distribution to test images and relatively accurate labels. Therefore, the web images with higher weights are more likely to be non-outliers and visually resembling test images. We take the Dogs dataset as an example and rank the web images based on the importance weight vector $\bm{\theta}$ learnt by our method. The web images with $5$ highest (\resp, lowest) weights are shown in the top (\resp, bottom) row in Fig.~\ref{fig:weights}, in which the numbers below images are their corresponding weights within the range $[0, 1.5]$ because the optimal upper bound of importance weights $b$ on the Dogs dataset obtained using cross-validation is $1.5$. From Fig.~\ref{fig:weights}, it can be seen that the top row of images with highest weights have accurate labels. Moreover, the dogs occupy the substantial center of the entire image and visually resemble the test images. In contrast, the web images in the bottom row are quite noisy. In detail, some images contain fake (\eg, printing) dogs or partially occluded dogs while some images are misled by the ambiguous category name or not relevant to the category name at all. 

\begin{figure}[h]
 	\centering
 	\begin{subfigure}[b]{0.1037\textwidth}
    		\includegraphics[width=1\textwidth]{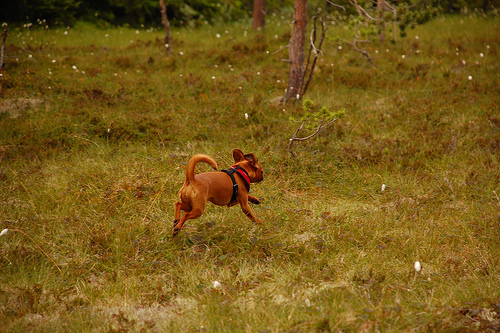}
		\caption{$1.46$}
    \end{subfigure}
    \begin{subfigure}[b]{0.0691\textwidth}
    		\includegraphics[width=1\textwidth]{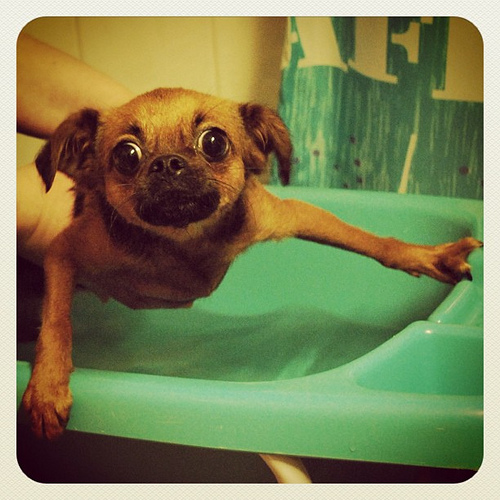}
		\caption{$1.46$}
    \end{subfigure}
    \begin{subfigure}[b]{0.0921\textwidth}
    		\includegraphics[width=1\textwidth]{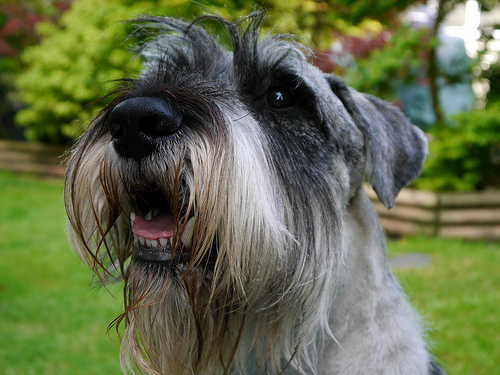}
		\caption{$1.24$}
    \end{subfigure}
    \begin{subfigure}[b]{0.1041\textwidth}
    		\includegraphics[width=1\textwidth]{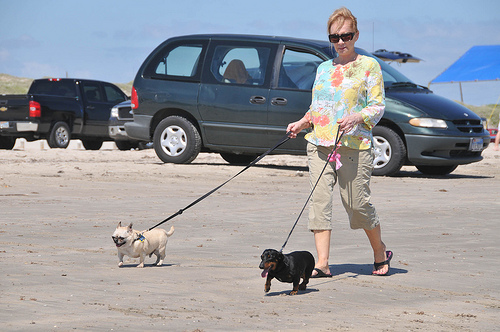}
		\caption{$1.23$}
    \end{subfigure}
    \begin{subfigure}[b]{0.0810\textwidth}
    		\includegraphics[width=1\textwidth]{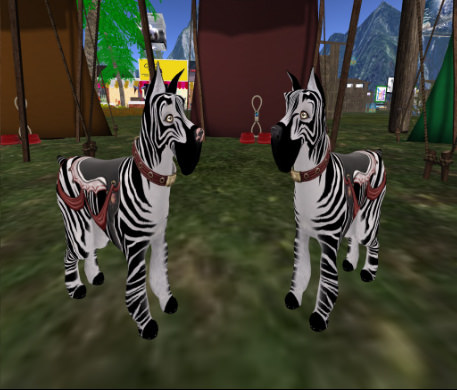}
		\caption{$1.20$}
    \end{subfigure}
    \caption{The web images associated with $5$ highest weights based on the weight vector $\bm{\theta}$ learnt by our Ours\_WSL.}
    \label{fig:more_weights}
\end{figure}

To demonstrate the superiority of our method compared with our special case Ours\_WSL in a qualitative fashion, we additionally show the web images associated with $5$ highest weights based on the weight vector $\bm{\theta}$ learnt by Ours\_WSL in Fig.~\ref{fig:more_weights}. By comparing Fig.~\ref{fig:more_weights} and Fig.~\ref{fig:weights}, we observe that the images in the top row in Fig.~\ref{fig:weights} have dominant centered objects while in the images in Fig.~\ref{fig:more_weights}, some objects are very small with cluttered background (\eg, (a) and (d)) or even not dogs (\eg, (e)), which indicates the advantage of using auxiliary categories to deal with the label noise of web images. 
We have similar observations on the other two datasets.

\begin{table}[t]
\caption{Accuracies (\%) of different methods on three datasets under the generalized setting. The best results are highlighted in boldface.}
\setlength{\tabcolsep}{5pt}
\label{tab:exp_generalized_results}
\centering
\begin{tabular}{|c|c|c|c|c|}
\hline
Dataset & CUB & SUN & Dogs & Avg \\
\hline
LR\_mix & 55.27 & 32.03 & 53.74 & 47.01 \\
WSL+LR & 57.60 & 35.11 & 55.13 & 49.28 \\
Chao~\etal~\cite{chao2016empirical} & 25.75 & 20.77 & 31.53 & 26.02\\
\hline
Ours & \textbf{59.60} & \textbf{36.00} & \textbf{65.89} & \textbf{53.83}\\
\hline
\end{tabular}
\end{table}

\noindent\textbf{Generalized ZSL Setting: }In some real-world applications, the test instances may come from both auxiliary categories and test categories. For instance, given the entire set of all fine-grained categories belonging to one coarse-grained category (\eg, 14,000 bird species), we annotate a few (\eg, 100) fine-grained categories and crawl web images for the remaining fine-grained categories. Then, we aim to recognize the test images from all fine-grained categories (\eg, 14,000). This generalized setting is more enticing yet more challenging. In fact, in the absence of web training images, this generalized setting reduces to the generalized Zero-Shot Learning (ZSL)~\cite{chao2016empirical}, in which the test instances may come from both seen and unseen categories. 

To investigate the effectiveness of our method under the generalized setting, we conduct extra experiments with the test instances from a mixture of auxiliary categories and test categories. Under the generalized setting, our method in (\ref{eqn:WSZSL1}) can be readily applied with a little abusively used dictionary $\D^t$ in $\|\X^t-\D^t\A^t\|_F^2$ for all the categories instead of only test categories. After the acquisition of $\A^t$, similarly as in Section~\ref{sec:method_web_image}, we adopt nearest neighbor (NN) classifier to predict test instances by comparing $\A^t$ with $\bar{\A}=[\bar{\A}^a, \bar{\A}^t]$, so that each test instance can be categorized into a test category or an auxiliary category. Following the setting in~\cite{chao2016empirical}, $20\%$ of the training instances in each auxiliary category are moved from training set to test set, leading to a new test set composed of the instances from both auxiliary and test categories. To be more specific, the new test set of CUB (\resp, SUN and Dogs) contains in total $200$ (\resp, $717$ and $113$) categories.

In terms of baselines, we compare our method with basic linear regression which learns a linear regressor for each auxiliary category and test category, which is referred to as LR\_mix in Table~\ref{tab:exp_generalized_results}. We also compare with WSL+LR which uses Xiao~\etal~\cite{xiao2015learning} for test categories and linear regressor for auxiliary categories, considering that Xiao~\etal~\cite{xiao2015learning} is the most competitive WSL baseline as reported in Table~\ref{tab:exp_results}. 
Moreover, we include generalized ZSL method in~\cite{chao2016empirical} as a baseline, which is specifically designed for the generalized ZSL setting~\cite{chao2016empirical}. 
The experimental results under the generalized setting are summarized in Table~\ref{tab:exp_generalized_results}. One observation is that the results drop sharply compared with those reported in Table~\ref{tab:exp_results}, because the generalized setting is a more challenging task with the test instances from both auxiliary categories and test categories. Nevertheless, our method still produces the best results on all three datasets, which indicates that our approach is still effective under the generalized setting. 

\subsection{Fine-grained Image Classification with Privileged Information} \label{sec:exp_PI}
In this section, the experimental setting is basically the same as that in Section~\ref{sec:exp_wo_PI}, that is, we use the same datasets and the same splits of auxiliary/test categories. 

\noindent\textbf{Textual Features: }For the web training images from test categories, we additionally extract textual features from the surrounding textual information of web images. Specifically, we crawl the tag, title, and comment of each web training image as raw textual information. Then, we build the vocabulary based on the top $2000$ most frequent words from the aggregated raw textual information, during which we perform stop-word removal to eliminate the meaningless words. Finally, we encode the textual information of each web image into a $2000$-dim term-frequency (TF) feature based on the vocabulary, leading to a $2000$-dim textual feature for each web image.

\noindent\textbf{Parameters: }By using privileged information, we introduce another trade-off parameter $\gamma$ in (\ref{eqn:WSZSL2_PI}), which is determined within the range $[10^{-3},10^{-2},\ldots,10^{3}]$ using the same cross-validation strategy as in Section~\ref{sec:exp_wo_PI} and the details are omitted here.

\noindent\textbf{Experimental Results: }We evaluate our method using privileged information and SVM+~\cite{LUPIparadigm}, which is trained based on web training images and their surrounding textual information. Similar to WSL+ZSL in Table~\ref{tab:exp_results}, we compare with a combo baseline WSDG\_PI+ZSL, which averages the test decision values from WSDG\_PI~\cite{niu2015visual} and ZSL baseline~\cite{zhang2016zero}. Note that WSDG\_PI~\cite{niu2015visual} can handle the label noise of web data and utilize privileged information at the same time.
Besides, we also include the results of LR, WSL+ZSL, and Ours from Table~\ref{tab:exp_results} for comparison.
The results of various approaches are reported in Table~\ref{tab:exp_results_PI}.

From Table~\ref{tab:exp_results_PI}, we can see that SVM+ (\resp, WSDG\_PI+ZSL) outperforms LR (\resp, WSL+ZSL), which shows that it is useful to utilize the additional textual information as privileged information. We also observe that after using privileged information, our approach achieves better performance on all three datasets, which shows the advantage of incorporating privileged information into our method for fine-grained image classification.

\begin{table}[t]
\caption{Accuracies (\%) of different methods with or without privileged information (PI) on three datasets. The best results are highlighted in boldface.}
\setlength{\tabcolsep}{5pt}
\label{tab:exp_results_PI}
\centering
\begin{tabular}{|c|c|c|c|c|}
\hline
Dataset & CUB & SUN & Dogs & Avg \\
\hline
LR  & 68.39 & 62.50 & 77.67 & 69.52\\
SVM+~\cite{LUPIparadigm} & 70.33 & 64.00 & 78.95 & 71.09\\
\hline
WSL+ZSL & 72.21 & 78.50 & 81.90 & 77.53 \\
WSDG\_PI~\cite{niu2015visual}+ZSL & 73.50  & 79.00 & 83.62 & 78.71\\
\hline
Ours & 76.47 & 84.50 & 85.16 & 82.04\\
Ours (with PI)  & \textbf{77.94} & \textbf{85.50} & \textbf{86.82} & \textbf{83.42}\\
\hline
\end{tabular}
\end{table}

\noindent\textbf{Generalized ZSL Setting: }To further investigate the effectiveness of privileged information, we also evaluate our method under the same generalized setting as in Section~\ref{sec:exp_wo_PI}, in which the test instances come from both auxiliary categories and test categories. Recall that in Section~\ref{sec:exp_wo_PI}, we compare with LR\_mix which learns a linear regressor for each auxiliary category and test category, and WSL+LR which uses Xiao~\etal~\cite{xiao2015learning} for test categories and linear regressor for auxiliary categories. Here, to take advantage of privileged information, we learn SVM+ for each test category and SVM for each auxiliary category (auxiliary categories do not have privileged information), leading to the baseline SVM+\_mix in Table~\ref{tab:exp_generalized_results_PI}. We also use WSDG\_PI~\cite{niu2015visual} for test categories and linear regressor for auxiliary categories, leading to the baseline WSDG\_PI+LR in Table~\ref{tab:exp_generalized_results_PI}.
Besides, we include the results of LR\_mix, WSL+LR, and Ours from Table~\ref{tab:exp_generalized_results} for comparison. 

Based on Table~\ref{tab:exp_generalized_results_PI}, we observe that SVM+\_mix (\resp, WSDG\_PI+LR) achieves better results than LR\_mix (\resp, WSL+LR), which indicates that it is helpful to utilize the additional textual information as privileged information under the generalized setting. Another observation is that our method is further improved by using privileged information and produces the best results on all three datasets, which again shows the benefit of privileged information under the generalized setting.

\begin{table}[t]
\caption{Accuracies (\%) of different methods with or without privileged information (PI) on three datasets under the generalized setting. The best results are highlighted in boldface.}
\setlength{\tabcolsep}{5pt}
\label{tab:exp_generalized_results_PI}
\centering
\begin{tabular}{|c|c|c|c|c|}
\hline
Dataset & CUB & SUN & Dogs & Avg \\
\hline
LR\_mix & 55.27 & 32.03 & 53.74 & 47.01 \\
SVM+\_mix & 56.26 & 33.58 & 54.53 & 48.12 \\
\hline
WSL+LR & 57.60 & 35.11 & 55.13 & 49.28 \\
WSDG\_PI~\cite{niu2015visual}+LR & 59.08 & 36.39 & 57.09 & 50.85\\
\hline
Ours & 59.60 & 36.00 & 65.89 & 53.83\\
Ours (with PI) & \textbf{60.82} & \textbf{37.84} & \textbf{67.29} & \textbf{55.32}\\
\hline
\end{tabular}
\end{table}

\section{Conclusion}
In this paper, a new learning scenario has been proposed for fine-grained image classification by using both web data and auxiliary categories. In this learning scenario, we have developed a method unifying webly supervised learning and zero-shot learning, which can transfer the knowledge from auxiliary categories to test categories and simultaneously handle the label noise and domain shift of web data. Moreover, our method has been further extended by taking advantage of the surrounding textual information of web images as privileged information. Comprehensive experiments have demonstrated the effectiveness of our proposed methods.

\appendix
\subsection{Solution to (\ref{eqn:WSZSL2_lag})} \label{sec:appendix_WSZSL}
For ease of representation, we rewrite the objective in (\ref{eqn:WSZSL2_lag}) as follows,
\begin{eqnarray} \label{eqn:WSZSL2_lag_copy}
\mathcal{L}_{\stackrel{\D^t,\A^t,\Z^t}{\E^w, \bm{\theta}\in\bm{\mathcal{S}}}} =\!\!\!\! &&\!\!\!\!\!\!\!\!\frac{1}{2}\|\X^t\!-\!\D^t\A^t\|_F^2 \!+\! \frac{\lambda_1}{2}\|\D^t\!-\!\D^a\|_F^2 \!+\!\lambda_2\|\Z^t\|_*\nonumber\\
&&\!\!\!\!\!\!\!\!\!\!\!\!\!\!\!\!\!\!\!\!\!\!\!\!\!\!\!\!\!\!\!\!\!\!\!\!\!\!+ \frac{\lambda_3}{2} \|\frac{1}{n^w}\X^w\bm{\theta}-\frac{1}{n^t}\X^t\1\|^2+ \lambda_4\|\E^w\|_{2,1}\nonumber\\
&&\!\!\!\!\!\!\!\!\!\!\!\!\!\!\!\!\!\!\!\!\!\!\!\!\!\!\!\!\!\!\!\!\!\!\!\!\!\!+\frac{\mu}{2}\|\E^w\!\!-\!(\X^w\!\!-\!\D^t\A^w)\bm{\Theta}\|_F^2\!+\!\left\langle\R,\!\E^w\!\!-\!(\X^w\!\!-\!\D^t\A^w)\bm{\Theta} \right\rangle\nonumber\\
&&\!\!\!\!\!\!\!\!\!\!\!\!\!\!\!\!\!\!\!\!\!\!\!\!\!\!\!\!\!\!\!\!\!\!\!\!\!\!+\frac{\mu}{2}\|\A^t-\Z^t\|_F^2+\left\langle\T, \A^t-\Z^t\right\rangle.
\end{eqnarray}

To minimize (\ref{eqn:WSZSL2_lag_copy}), we update $\E^w, \Z^t$, $\D^t$, $\A^t$, and $\bm{\theta}$ one by one in an alternating fashion, which will be detailed in the following.

\setlength{\textfloatsep}{5pt}
\begin{algorithm}[t]
   \caption{Solving (\ref{eqn:WSZSL2}) with inexact ALM}
   \label{alg:WZSL}
\begin{algorithmic}[1]
   \STATE {\bfseries Input:} $\X^a, \A^a, \X^w, \A^w, \X^t, \D^a$.
   \STATE Initialize $\R=\O$, $\T=\O$, $\bm{\theta}=\1$, $\D^t=\D^a$, $\rho = 0.1$, $\mu = 0.1$, $\mu_{max}=10^6$, $\nu=10^{-5}$, $N_{iter}=10^6$.
   \FOR{ $t = 1 : N_{iter}$ }
      \STATE Update $\E^w$ by using (\ref{eqn:ea_sol}).
      \STATE Update $\Z^t$ by using (\ref{eqn:zt_sol}).
	  \STATE Update $\D^t$ by using (\ref{eqn:dt_sol}).      
      \STATE Update $\A^t$ by using (\ref{eqn:at_sol}).     
      \STATE Update $\bm{\theta}$ by solving  (\ref{eqn:bt3}).
      \STATE Update $\R$ by $\R = \R+\mu(\E^w-(\X^w-\D^t\A^w)\bm{\Theta})$.
      \STATE Update $\T$ by $\T = \T+\mu(\A^t-\Z^t)$.
      \STATE Update the parameter $\mu$ by $\mu\!=\!\min(\mu_{max},(1\!+\!\rho)\mu)$.    
      \STATE Break if $\|\E^w-(\X^w-\D^t\A^w)\bm{\Theta}\|_\infty<\nu$ and $\|\A^t-\Z^t\|_\infty<\nu$. 
   \ENDFOR
   \STATE {\bfseries Output:} $\A^t$.
\end{algorithmic}
\label{alg:RKLRR}
\end{algorithm}

\noindent\textbf{Update $\E^w$: }The subproblem of (\ref{eqn:WSZSL2_lag_copy}) \wrt $\E^w$ is as follows,
\begin{eqnarray} \label{eqn:ea}
\min_{\E^w}\lambda_4\|\E^w\|_{2,1} + \frac{\mu}{2}\|\E^w- \left((\X^w-\D^t\A^w)\bm{\Theta}-\frac{\R}{\mu}\right)\|_F^2,\nonumber
\end{eqnarray}
which has a closed-form solution~\cite{liu2010robust}. Specifically, by denoting $\Q=(\X^w-\D^t\A^w)\bm{\Theta}-\frac{\R}{\mu}$, if the optimal solution \wrt $\E^w$ is $\E^{w*}$, then the $i$-th column of $\E^{w*}$ is
\begin{eqnarray} \label{eqn:ea_sol}
\E^{w*}(:,i)=
\begin{cases}
\frac{\|\q_i\|_2-\frac{\lambda_4}{\mu}}{\|\q_i\|_2}\q_i,  \quad & \textnormal{if}\,\, \frac{\lambda_4}{\mu}<\|\q_i\|_2, \\
0, \quad &\textnormal{otherwise},
\end{cases}
\end{eqnarray}
\noindent where $\q_i$ is the $i$-th column of $\Q$ and $\|\q_i\|_2$ is the $L_2$ norm of $\q_i$.

\noindent\textbf{Update $\Z^t$: }
The subproblem of (\ref{eqn:WSZSL2_lag_copy}) \wrt $\Z^t$ is as follows,
\begin{eqnarray} \label{eqn:zt}
\min_{\Z^t} \lambda_2\|\Z^t\|_* + \frac{\mu}{2}\|\Z^t-(\A^t+\frac{\T}{\mu})\|_F^2,
\end{eqnarray}
\noindent which can be solved based on Singular Value Threshold (SVT) method \cite{cai2010singular}. By denoting $\M=\A^t+\frac{\T}{\mu}$ and the rank of $\M$ as $r$, the singular value decomposition of $\M$ can be represented as $\M = \U\bm{\Sigma}\V'$, where $\U\in\mathcal{R}^{m\times r},\V\in\mathcal{R}^{r\times n^t}$, and $\bm{\Sigma}=\bbR^{r\times r}$ is a diagonal matrix with diagonal entries being the singular values of $\M$. Then, the solution \wrt $\Z^t$ can be obtained as follows,
\begin{eqnarray} \label{eqn:zt_sol}
\Z^t=\U\mathcal{D}(\bm{\Sigma})\V',
\end{eqnarray}
\noindent where $\mathcal{D}(\bm{\Sigma})$ is a diagonal matrix with   $\{(\sigma_i-\frac{\lambda_2}{\mu})_{+}|_{i=1}^r\}$ being the diagonal elements, in which $\sigma_i$ is the $i$-th diagonal entry of $\bm{\Sigma}$ and $(\cdot)_{+}$ is an operator setting the negative entries to zeros.

\noindent\textbf{Update $\D^t$: }
The subproblem of (\ref{eqn:WSZSL2_lag_copy}) \wrt $\D^t$ is as follows,

\noindent
\begin{eqnarray} \label{eqn:dt}
\min_{\D^t} &&\frac{1}{2}\|\X^t-\D^t\A^t\|_F^2 + \frac{\lambda_1}{2}\|\D^t-\D^a\|_F^2\\
&&\!\!\!\!\!\!\!\!\!\!\!\!\!\!\!\!\!\!\!\!\!\!\!\!\!+ \frac{\mu}{2}\|\E^w\!\!-\!(\X^w\!\!-\!\D^t\A^w)\bm{\Theta}\|_F^2\!+\!\left\langle\R,\!\E^w\!\!-\!(\X^w\!\!-\!\D^t\A^w)\bm{\Theta} \right\rangle.\nonumber
\end{eqnarray}

\noindent We set the derivative of (\ref{eqn:dt}) \wrt $\D^t$ as zeros, and obtain the closed-form solution to $\D^t$ as
\begin{eqnarray} \label{eqn:dt_sol}
\D^t = \!\!\!\!\!\!\!\!\!\!&&\left(\X^t{\A^t}'+\lambda_1\D^a+(\mu\X^w\bm{\Theta}-\mu\E^w-\R)\bm{\Theta}'{\A^w}'\right)\nonumber\\
&&\left(\A^t{\A^t}'
+\mu\A^w\bm{\Theta}\bm{\Theta}'{\A^w}'+\lambda_1\I\right)^{-1}
.\end{eqnarray}

\noindent\textbf{Update $\A^t$: }
The subproblem of (\ref{eqn:WSZSL2_lag_copy}) \wrt $\A^t$ is as follows,
\begin{eqnarray}\label{eqn:at}
\min_{\A^t}\frac{1}{2}\|\X^t-\D^t\A^t\|_F^2 +\frac{\mu}{2}\|\A^t-\Z^t\|_F^2+\left\langle\T, \A^t-\Z^t\right\rangle,\nonumber
\end{eqnarray}
which also has a closed-form solution:
\begin{eqnarray} \label{eqn:at_sol}
\A^t=\left({\D^t}'\D^t+\mu\I\right)^{-1}({\D^t}'\X^t+\mu\Z^t-\T).
\end{eqnarray}

\noindent\textbf{Update $\bm{\theta}$: }The subproblem of (\ref{eqn:WSZSL2_lag_copy}) \wrt $\bm{\theta}$ is as follows,
\begin{eqnarray} \label{eqn:bt}
\!\!\!\!\!\!\!\!\!\!\!\!\!\!\!\!\min_{\bm{\theta}\in \bm{\mathcal{S}}}\!\!\!\!\!\!\!\!\!\!\!\!&&\frac{\lambda_3}{2} \|\frac{1}{n^w}\X^w\bm{\theta}\!-\!\frac{1}{n^t}\X^t\1\|^2\!+\!\frac{\mu}{2}\|\E^w\!\!-\!(\X^w\!\!-\!\D^t\A^w)\bm{\Theta}\|_F^2  \nonumber\\
&&+ \left\langle\R, \E^w-(\X^w-\D^t\A^w)\bm{\Theta} \right\rangle.
\end{eqnarray}

\noindent After omitting the constant terms without $\bm{\theta}$, the problem in (\ref{eqn:bt}) can be converted to
\begin{eqnarray}  \label{eqn:bt2}
\min_{\bm{\theta}\in\bm{\mathcal{S}}}&&\frac{\lambda_3}{2{(n^w)}^2}\bm{\theta}'{\X^w}'\X^w\bm{\theta}-\frac{\lambda_3}{n^w n^t}\bm{\theta}'{\X^w}'\X^t\1
\nonumber\\
&&+ \frac{\mu}{2} \bm{\theta}'\bar{\P}\bm{\theta} - \mu\bm{\theta}'\hat{\p} - \bm{\theta}'\hat{\r},
\end{eqnarray}

\noindent in which $\bar{\P}$ is a diagonal matrix sharing the same diagonal with $(\X^w-\D^t\A^w)'(\X^w-\D^t\A^w)$, $\hat{\p}$ is the diagonal of $(\X^w-\D^t\A^w)'\E^w$, and $\hat{\r}=(\R\circ(\X^w-\D^t\A^w))'\1$. The problem in (\ref{eqn:bt2}) can be further simplified as
\begin{eqnarray} \label{eqn:bt3}
\min_{\bm{\theta}\in\bm{\mathcal{S}}} \frac{1}{2} \bm{\theta}'\H\bm{\theta} - \f'\bm{\theta},
\end{eqnarray}
\noindent in which $\H=\frac{\lambda_3}{{(n^w)}^2}{\X^w}'\X^w+\mu\bar{\P}$ and $\f=\frac{\lambda_3}{n^w n^t}{\X^w}'\X^t\1$ $+$ $\mu\hat{\p}$ $+$ $\hat{\r}$. The problem in (\ref{eqn:bt3}) is known as quadratic programming (QP) problem and could be solved using off-the-shelf QP solvers (\eg, Mosek). However, based on our experimental observation, existing QP solvers are not very efficient. So we develop our own Sequential Minimal Optimization (SMO)~\cite{platt1998sequential} based algorithm to solve (\ref{eqn:bt3}), which is much more efficient than those off-the-shelf QP solvers. Simply speaking, we select the most violating pair of variables in $\bm{\theta}$ for update in each iteration sequentially until the objective of (\ref{eqn:bt3}) converges. 

The whole algorithm using inexact ALM is summarized in Algorithm~\ref{alg:WZSL}. Based on our experimental observation, the algorithm usually converges within $50$ iterations.

\subsection{Solution to (\ref{eqn:WSZSL2_PI})} \label{sec:appendix_WSZSL_PI}
The problem in (\ref{eqn:WSZSL2_PI}) can be solved similarly to (\ref{eqn:WSZSL2}). We write the  augmented Lagrangian function of (\ref{eqn:WSZSL2_PI}) as follows,
\begin{eqnarray} \label{eqn:WSZSL2_PI_lag}
\mathcal{L}_{\stackrel{\D^t,\A^t,\Z^t}{\E^w, \bm{\theta}, \tilde{\W}}} =\!\!\!\! &&\!\!\!\!\!\!\!\!\frac{1}{2}\|\X^t\!-\!\D^t\A^t\|_F^2 \!+\! \frac{\lambda_1}{2}\|\D^t\!-\!\D^a\|_F^2 \!+\!\lambda_2\|\Z^t\|_*\nonumber\\
&&\!\!\!\!\!\!\!\!\!\!\!\!\!\!\!\!\!\!\!\!\!\!\!\!\!\!\!\!\!\!\!\!\!\!\!\!\!\!+ \frac{\lambda_3}{2} \|\frac{1}{n^w}\X^w\bm{\theta}-\frac{1}{n^t}\X^t\1\|^2+ \lambda_4\|\E^w\|_{2,1}\nonumber\\
&&\!\!\!\!\!\!\!\!\!\!\!\!\!\!\!\!\!\!\!\!\!\!\!\!\!\!\!\!\!\!\!\!\!\!\!\!\!\! +\frac{\gamma}{2}\|(\X^w-\D^t\A^w)-\tilde{\W}\tilde{\X}^w\|_F^2\nonumber\\
&&\!\!\!\!\!\!\!\!\!\!\!\!\!\!\!\!\!\!\!\!\!\!\!\!\!\!\!\!\!\!\!\!\!\!\!\!\!\!+\frac{\mu}{2}\|\E^w\!\!-\!(\X^w\!\!-\!\D^t\A^w)\bm{\Theta}\|_F^2\!+\!\left\langle\R,\!\E^w\!\!-\!(\X^w\!\!-\!\D^t\A^w)\bm{\Theta} \right\rangle\nonumber\\
&&\!\!\!\!\!\!\!\!\!\!\!\!\!\!\!\!\!\!\!\!\!\!\!\!\!\!\!\!\!\!\!\!\!\!\!\!\!\!+\frac{\mu}{2}\|\A^t-\Z^t\|_F^2+\left\langle\T, \A^t-\Z^t\right\rangle, 
\end{eqnarray}
which can be minimized  by updating $\E^w, \Z^t$, $\D^t$, $\A^t$, $\bm{\theta}$, and $\tilde{\W}$ one by one iteratively until the termination criterion is met. Compared with the solution to (\ref{eqn:WSZSL2}), the only difference lies in the steps of updating $\D^t$ and $\tilde{\W}$, which will be elaborated below.

\noindent\textbf{Update $\D^t$: }
The subproblem of (\ref{eqn:WSZSL2_PI_lag}) \wrt $\D^t$ is as follows,
\begin{eqnarray} \label{eqn:dt_PI}
\min_{\D^t} &&\frac{1}{2}\|\X^t-\D^t\A^t\|_F^2 + \frac{\lambda_1}{2}\|\D^t-\D^a\|_F^2\\
&&\!\!\!\!\!\!\!\!\!\!\!\!\!\!\!\!\!\!\!\!\!\!\!\!\!+ \frac{\mu}{2}\|\E^w\!\!-\!(\X^w\!\!-\!\D^t\A^w)\bm{\Theta}\|_F^2\!+\!\left\langle\R,\!\E^w\!\!-\!(\X^w\!\!-\!\D^t\A^w)\bm{\Theta} \right\rangle\nonumber\\
&&\!\!\!\!\!\!\!\!\!\!\!\!\!\!\!\!\!\!\!\!\!\!\!\!\!+\frac{\gamma}{2}\|(\X^w-\D^t\A^w)-\tilde{\W}\tilde{\X}^w\|_F^2.
\nonumber
\end{eqnarray}

\noindent We set the derivative of (\ref{eqn:dt_PI}) \wrt $\D^t$ as zeros and arrive at the following closed-form solution to $\D^t$:
\begin{eqnarray} \label{eqn:dt_PI_sol}
\D^t = \!\!\!\!\!\!\!\!\!\!&&[\X^t{\A^t}'+\lambda_1\D^a+(\mu\X^w\bm{\Theta}-\mu\E^w-\R)\bm{\Theta}'{\A^w}' \nonumber\\
&&+\gamma \X^w -\gamma\tilde{\W}\tilde{\X}^w\A^{w'}]\nonumber\\
&&\!\!\!\!\left(\A^t{\A^t}'
+\mu\A^w\bm{\Theta}\bm{\Theta}'{\A^w}'+\lambda_1\I+\gamma\A^w\A^{w'}\right)^{-1}\!\!\!\!.
\end{eqnarray}

\noindent\textbf{Update $\tilde{\W}$: }
The subproblem of (\ref{eqn:WSZSL2_PI_lag}) \wrt $\tilde{\W}$ is as follows,
\begin{eqnarray} \label{eqn:dtW_PI}
\min_{\tilde{\W}} \frac{\gamma}{2}\|(\X^w-\D^t\A^w)-\tilde{\W}\tilde{\X}^w\|_F^2.
\end{eqnarray}
\noindent We set the derivative of (\ref{eqn:dtW_PI}) \wrt $\tilde{\W}$ as zeros, and obtain the closed-form solution to $\tilde{\W}$ as
\begin{eqnarray} \label{eqn:dtW_PI_sol}
\tilde{\W} =(\X^w\tilde{\X}^{w'}-\D^t\A^w\tilde{\X}^{w'})(\tilde{\X}^w\tilde{\X}^{w'})^{-1}
.\end{eqnarray}

\bibliographystyle{IEEEtran}
\bibliography{main}

\end{document}